\documentclass[sigconf]{acmart}

\usepackage{subcaption}
\usepackage{multirow}
\usepackage{pdflscape}
\usepackage{makecell}
\usepackage{amsmath}

\usepackage{array}

\usepackage{colortbl}

\usepackage[title]{appendix}

\copyrightyear{2025}
\acmYear{2025}
\setcopyright{cc}
\setcctype{by}
\acmConference[FDG '25]{International Conference on the Foundations of Digital Games}{April 15--18, 2025}{Graz, Austria}
\acmBooktitle{International Conference on the Foundations of Digital Games (FDG '25), April 15--18, 2025, Graz, Austria}
\acmPrice{}
\acmDOI{10.1145/3723498.3723794}
\acmISBN{/25/04}

\begin{document}

\title{The Procedural Content Generation Benchmark:\\
An Open-source Testbed for Generative Challenges in Games}

\author{Ahmed Khalifa}
\affiliation{%
  \institution{Institute of Digital Games}
  \institution{University of Malta}
  \city{Msida}
  \country{Malta}
}
\email{ahmed@akhalifa.com}

\author{Roberto Gallotta}
\affiliation{%
  \institution{Institute of Digital Games}
  \institution{University of Malta}
  \city{Msida}
  \country{Malta}
}
\email{roberto.gallotta@um.edu.mt}

\author{Matthew Barthet}
\affiliation{%
  \institution{Institute of Digital Games}
  \institution{University of Malta}
  \city{Msida}
  \country{Malta}
}
\email{matthew.barthet@um.edu.mt}

\author{Antonios Liapis}
\affiliation{%
   \institution{Institute of Digital Games}
  \institution{University of Malta}
  \city{Msida}
  \country{Malta}
}
\email{antonios.liapis@um.edu.mt}

\author{Julian Togelius}
\affiliation{%
  \institution{Game Innovation Lab}
  \institution{New York University}
  \city{New York}
  \state{New York}
  \country{USA}
}
\email{julian@togelius.com}

\author{Georgios N. Yannakakis}
\affiliation{%
  \institution{Institute of Digital Games}
  \institution{University of Malta}
  \city{Msida}
  \country{Malta}
}
\email{georgios.yannakakis@um.edu.mt}

\renewcommand{\shortauthors}{Khalifa et al.}

\begin{abstract}
This paper introduces the Procedural Content Generation Benchmark for evaluating generative algorithms on different game content creation tasks. The benchmark comes with 12 game-related problems with multiple variants on each problem. Problems vary from creating levels of different kinds to creating rule sets for simple arcade games. Each problem has its own content representation, control parameters, and evaluation metrics for quality, diversity, and controllability. This benchmark is intended as a first step towards a standardized way of comparing generative algorithms. We use the benchmark to score three baseline algorithms: a random generator, an evolution strategy, and a genetic algorithm. Results show that some problems are easier to solve than others, as well as the impact the chosen objective has on quality, diversity, and controllability of the generated artifacts.
\end{abstract}



\maketitle

\section{Introduction}
\label{sec:introduction}

Scientific and technological progress requires reliable means of measuring the performance of methods and machines. Therefore, extensive efforts focus on developing tools and techniques for such measurements within various scientific and engineering fields. In Artificial Intelligence (AI), \textit{benchmarks} are crucial to the field: most AI papers include a comparison with the state of the art on some benchmark. In Reinforcement Learning (RL) research, \textit{Montezuma's Revenge} (Utopia Software, 1984) and \textit{Pitfall!} (Activision, 1984) were two games---among 100 games in the Arcade Learning Environment~\cite{bellemare_2013_arcade}---which showcased the problems with deep RL approaches and pushed researchers to innovate~\cite{ecoffet2021first}. However, some qualities are easier to benchmark than others. While an algorithm's speed and memory footprint are easily measurable, qualities related to creative expression are inherently complicated to measure---especially via automated means. This does not make such benchmarks less important \cite{guzdial2019benchmark}.

While it may be hard to directly measure the creativity of the output of a software system \cite{ritchie2007empirical}, one can measure the usefulness of software that forms part of a creative system. Good measures of usefulness would assess how well the output performs according to some functionality measure, how diverse the output is, and how well the generator can be controlled. Having quantifiable measures of usefulness allows AI researchers to identify and address limitations in current algorithms. There are currently very few benchmarks in Procedural Content Generation (PCG) and generative AI in games (see Section~\ref{sec:relatedwork_benchmarks}). A one-stop framework that researchers and newcomers can use to explore the generative problems in games can lead to more rigorous testing protocols that can change the frontier of PCG research---in the same way that Arcade Learning Environment changed RL research.

In this paper, we introduce the Procedural Content Generation Benchmark (PCG Benchmark) which marks the first step towards standardizing problems in the generative space in games. It offers a set of problems where researchers can explore different methods and understand what works and what does not, as well as an extensible framework for adding further PCG problems. The PCG Benchmark comes with 12 problems out of the box, including the generation of game rules, levels, buildings, word games, and patterns (see Section \ref{sec:protocol_problems}). Every problem has to follow the same evaluation criteria, using an array of content as input and returning the success score in terms of quality, diversity, and controllability as output (see Section~\ref{sec:pcg_benchmark}). Reaching the maximum score on these problems does not necessitate that challenges in generating content for this game have all been overcome. It just means that this particular formulation of the problem with these particular criteria is solved. We test the framework against three baseline algorithms that follow the search-based PCG paradigm~\cite{togelius_2011_search}; results showcase the different challenges posed by the multi-faceted generative problems already implemented into the PCG Benchmark.

\section{Background}
This section takes a brief look at the history of PCG in games and the benchmarks that have advanced Game AI research.

\subsection{Procedural Content Generation in Games}
\label{sec:relatedwork_pcg}

PCG is central to technical games research. Early applications of PCG in games focused on adding scale and replayability to games with limited hardware, such as the infinite level layouts of \emph{Rogue} (Toy and Wichman, 1980), or the large star systems provided in \emph{Elite} (Acornsoft, 1984).  In \emph{search-based} PCG \cite{togelius_2011_search}, artificial evolution or similar stochastic optimization methods are used to generate content that optimizes an evaluation function. Search-based PCG provides the capacity to generate more complex content while retaining functionality guarantees. Some notable examples of search-based PCG include generating weapons for \emph{Galactic Arms Race} \cite{hastings_2009_automatic}, generating new level layouts for \emph{Super Mario Bros.} (Nintendo, 1985) \cite{shaker2012evolving}, and generating novel suggestions as a designer assistant tool \cite{liapis2013sentient, charity_2020_baba}. Experience-driven PCG \cite{yannakakis_2011_experience} combines the search-based approach with player models to generate content that elicits a desired experience in the player, such as \emph{Super Mario Bros} levels~\cite{wang_2022_fun} which maximize a diversity metric based on Raph Koster's \emph{theory of fun}~\cite{koster_2013_theory}, or levels which elicit emotional trajectories during play \cite{lopes_2015_targeting}. 

Within search-based PCG, quality diversity (QD) evolutionary algorithms \cite{pugh2016qd} can be used to generate a set of high-quality solutions across a range of behavior metrics to ensure meaningful diversity in the output~\cite{gravina_2019_procedural}. PCG through QD has become a popular method for generators focused on creativity, such as for generating novel game levels \cite{beukman_2022_procedural}, bullet patterns~\cite{khalifa2018talakat}, interesting 2D \cite{liapis_2013_transforming,liapis2016arcade} and 3D \cite{gallotta2023preference} spaceships, and diverse \emph{Minecraft} (Mojang, 2011) buildings \cite{barthet_2022_open}.

Perhaps the biggest downside of search-based PCG is the computational cost of content production. PCG via machine learning (PCGML) \cite{summerville_2018_procedural} instead moves the computational cost to the training phase, with generally much faster inference. Many examples of PCGML in games leverage self-supervised learning on existing game content to learn to generate game levels, e.g. using generative adversarial networks \cite{volz_2018_evolving}, wave function collapse \cite{kim_2019_automatic}, or computer vision \cite{guzdial_2016_toward}. But this requires a sufficient amount of game content to train on. PCG via reinforcement learning (PCGRL) \cite{khalifa_2020_pcgrl} instead trains RL agents to generate new content based only on rewards, and can generate new content in an online fashion \cite{wang_2022_fun}.  

Looking at the evolution of PCG over time~\cite{liapis2020tenyears}, we are always pushing towards using new and novel systems to generate content. Establishing a standardized and open-sourced approach for evaluating the capabilities of these algorithms ensures transparency and replicability of results~\cite{bodria_2023_benchmarking}. With the rise of new AI methods~\cite{gallotta2024large}, having a comprehensive way to compare large language models or other generative AI methods to earlier algorithms is crucial. Importantly, an easy-to-use benchmark can act as a teaching tool for PCG at an undergraduate or graduate level, allowing students to compare their algorithms' performance.

\subsection{Game AI Benchmarks}
\label{sec:relatedwork_benchmarks}

AI has a long-standing history of using game-based benchmarks as a point of comparison between new methods. \emph{Chess} and \emph{Go} famously played a significant role in the development of AI with the development of \emph{DeepBlue} \cite{campbell_2002_deep} and \emph{AlphaGo} \cite{silver_2016_mastering} respectively. The field eventually transitioned to tackle the complexity of digital games such as the \emph{Mario AI} benchmark \cite{karakovskiy_2012_mario} and \emph{Starcraft II} (Blizzard, 2011) \cite{vinyals_2019_grandmaster}. Training gameplaying agents via RL \cite{mnih_2013_playing} has relied on standardized and readily available benchmarks built on \textit{OpenAI Gym} \cite{brockman_2016_openai} as an easy point of comparison between methods. A notable example is the \emph{Arcade Learning Environment} \cite{bellemare_2013_arcade}, which consists of over 100 game environments for the \emph{Atari 2600} console spanning a variety of game genres. The development of these benchmarks was pivotal towards the advancement of gameplaying AI~\cite{mnih_2013_playing}. Newer gameplaying benchmarks even include a player experience component \cite{barthet2024affectively}, targeting more believable and human-like play.

With the rise of generative AI, similar benchmarks are needed to push toward better methods that can work for games. The history of PCG research already includes various PCG testbeds and competitions, such as competitions on \emph{Super Mario Bros.} (Nintendo, 1985) \cite{horn_2014_comparative}, Generative Design in \emph{Minecraft} (GDMC) \cite{salge_2018_generative, grbic_2021_evocraft}, and the level and rule generation tracks in GVGAI \cite{perez_2019_general}. However, there are currently no unified benchmarks for PCG which encompass a variety of problems and facets of content generation. One critical difference between the benchmarks for PCG and gameplaying is that evaluating the output of creative systems is often ill-defined, complicated to measure, and subjective. For example, the GDMC~\cite{salge_2018_generative} competition employs human judges to evaluate the submitted algorithms. This human-centered evaluation cannot scale well and can be influenced by biases \cite{yannakakis_2014_panorama}. Evaluating the output of generated content is usually problem-specific due to the huge variety in content types and representations. Through our benchmark, we aim to provide a standardized platform for an easier comparison of generators across a common set of problems.

\section{PCG Benchmark}
\label{sec:pcg_benchmark}

The PCG Benchmark\footnote{\url{https://github.com/amidos2006/pcg_benchmark}} is an easy-to-use framework that allows users to evaluate their generative algorithm against a multitude of generative problems in games. The framework follows the design concepts of \textit{OpenAI Gym}~\cite{brockman_2016_openai}: each problem is independent and has its own representation and evaluation criteria. The framework provides an evaluation function that can be used to evaluate any content on three criteria:
\begin{itemize}
\item \textbf{Quality} measures the percentage of the input content that passes the quality criteria for the current problem. For example, if the problem is generating levels for Mario with quality criteria of having playable levels and the user provided 100 levels to evaluate, the system will test all the levels, and if 20 levels are playable then the result will be 20\%.
\item \textbf{Diversity} measures the percentage of the input content that passes the diversity criteria for the current problem. This is an important aspect in evaluating any PCG algorithm as having a generator that generates the same content or small variants of the same content every time should be considered a bad generator~\cite{compton2015casual}. For example, if the problem is generating full games, having a generator providing 100 games where all of them are variants of a block-pushing game (a-la \textit{Sokoban}) with just different named objects then diversity should be low.
\item \textbf{Controllability} measures the percentage of the input content that adheres to some controllability constraints. Controllability has to be tested against a control parameter that the user should provide; such parameters are designer choices (e.g. the number of enemies in a generated level), rather than a playability constraint. 
Without having control parameters paired with the input content, controllability can not be measured; the system will always return 0\%. Not every generator needs to be controllable: developers can always create a new generator for every different problem. However, having a generator that adapts to target parameters without changing any of the underlying code makes it easier to use during production.
\end{itemize}

Besides these percentages, the benchmark returns additional details about every artifact in the input. Each artifact receives three values within $[0,1]$ that capture how close it is to passing each criterion (quality, diversity, or controllability). This can be used as an error function for PCGML, as a reward for PCGRL, or as a fitness function for search-based PCG (see Section \ref{sec:relatedwork_pcg}). Additional, problem-specific results can also be provided per artifact. For example, in the \textit{Zelda} problem (see Section \ref{sec:protocol_problems}), the solution length and level connectivity are part of the additional information that is returned per generated level in the input. This additional information can help the user to create more complex generators such as Quality Diversity algorithms~\cite{gravina_2019_procedural}, using these metrics as behavior characterizations~\cite{mouret2015illuminating} for the generated content.

The framework also provides two possibility spaces~\cite{cook2019generative}, one for the content and one for the control parameters. These spaces are designed to be similar to the \textit{OpenAI Gym} spaces for observations and actions~\cite{brockman_2016_openai}. Our spaces, however, define the possible values for both content and control parameters. These two spaces can be used to sample randomly from, validate if a content or a control parameter is possible, mix two artifacts probabilistically, change parts of an artifact randomly, convert an artifact to a flat (string) representation and back, etc. These functionalities allow users to build their generators with less friction. For example, a random generator just samples randomly from the content space and can keep the best-discovered artifacts. Moreover, access to these spaces allows the framework to have different representations for each generative problem: the user must then build their generator to work with each representation. This allows the system to work with any type of content in games such as text, level generation, rule generation, patterns, etc. Besides these functions, the framework has a render function that converts the content into the corresponding graphical representation. This graphical representation depends on the problem at hand: images, videos, strings, sounds, etc.

\begin{figure}
    \centering
    \includegraphics[width=\linewidth]{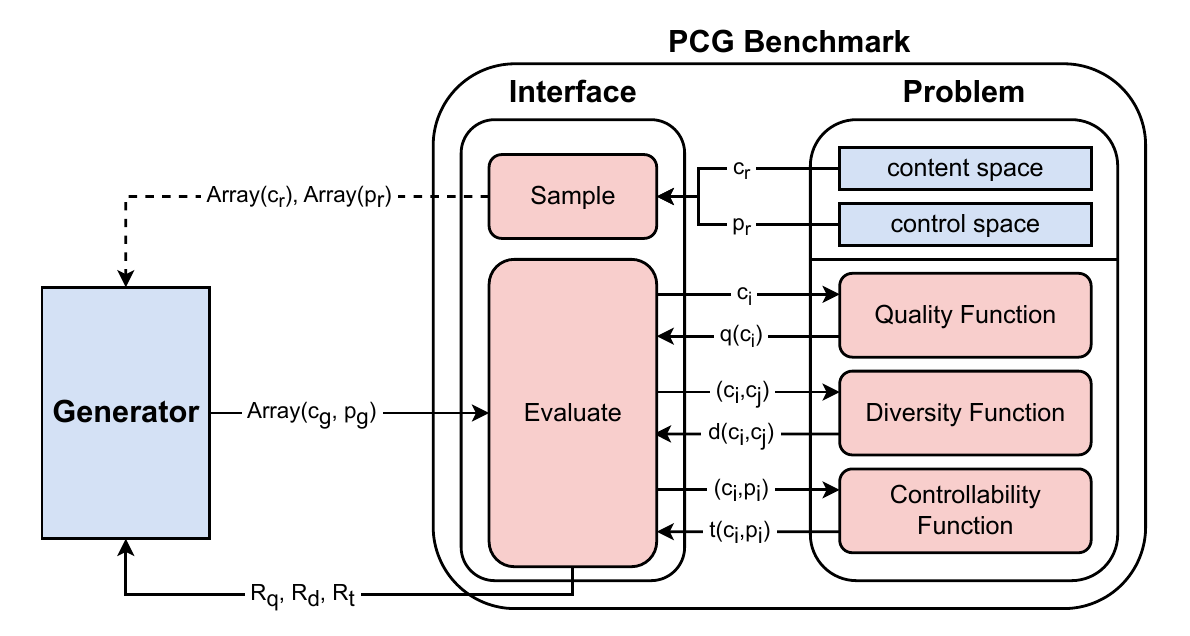}
    \caption{The system diagram for the PCG Benchmark that showcases how to use the framework. First, the \textbf{Generator} can sample an array of random content ($Array(c_r)$) and an array of random control parameters ($Array(p_r)$). Then the \textbf{Generator} returns an array of paired content and control parameters ($Array(c_g, p_g)$) to be evaluated. The system sends them to the current problem to calculate their values ($q(c_i)$ is the quality value for a content, $d(c_i,c_j)$ is the diversity value between two content, and $t(c_i,p_i)$ is the controllability value between a content and a control parameter). Finally, the system returns the results for quality ($R_q$), diversity ($R_d$), and controllability ($R_t$).}
    \label{fig:pcg_benchmark}
\end{figure}

\subsection{Using the Benchmark}

Figure~\ref{fig:pcg_benchmark} shows the system diagram of the PCG Benchmark and how a \textbf{Generator} can interact with all the moving parts through a simple interface. The user needs first to specify the problem that they need to solve (see section~\ref{sec:protocol_problems}). After that, the user can build their generator using any method, including constructive~\cite{shaker2016constructive}, search-based~\cite{togelius_2011_search}, quality diversity~\cite{gravina_2019_procedural}, machine learning~\cite{summerville_2018_procedural}, RL-based~\cite{khalifa_2020_pcgrl}, constrained-based~\cite{karth2017wavefunctioncollapse}, and other methods. The generator can use the control space, content space, and the evaluate function as many times as needed, even integrating them in its internal generative loop, e.g. in search-based methods \cite{togelius_2011_search}. Finally, the generator will return the generated content paired with the control parameters that this content was generated for\footnote{If values for the controllability parameters are not provided by the user, the algorithm returns the generated content only.}, and the benchmark will evaluate the content and return the scores for quality, diversity, and controllability (if applicable).

\subsection{Extending the Benchmark}
To make the framework easy to extend, we separated the interface where the generator interacts from the actual generative problem. When a user wants to test their algorithm, they ask the system for a specific problem using its unique name. The system finds it from the list of all the registered problems and returns it inside the fixed environment interface. Figure~\ref{fig:pcg_benchmark} shows a constructed environment for a specific problem. Due to this separation, adding new problems is easy. The user needs to create a new problem class that has the following functions:
\begin{itemize}
    \item \textbf{Info Function} takes as input the artifact and outputs an object that contains all the needed and related information of that content. This function is crucial, as it is used by most other functions (below). For example, if we are evaluating mazes, we might need to have the maze solution precomputed for quality (have a maze with at least 20 steps to solve), diversity (mazes with different solutions), and controllability (have a maze with exactly $X$ steps to solve where $X$ is the control parameter).
    \item \textbf{Quality Function} takes the info object and returns a value within $[0,1]$ that reflects how close that content is to passing the quality criteria (at $1.0$ it passes). In our maze generation example, this could be how close the solution length is to 20 or more.
    \item \textbf{Diversity Function} takes as input two info objects for different artifacts and outputs a value within $[0,1]$ that reflects how similar these two artifacts are to each other ($1.0$ means they are different and $0.0$ means they are identical). In our maze generation example, diversity can be on the actions taken for solving the maze, and if the solutions are at least 5 actions different between the two mazes then diversity is 1.0.
    \item \textbf{Controllability Function} takes as input the info object and value(s) for its control parameter(s) and outputs a value within $[0,1]$ that reflects how close the content is to match these values. If a control parameter is the number of enemies in a generated level, the function will return {1.0} if the generated has a number of enemies within an allowed range of this parameter's values.
    \item \textbf{Render Function} takes as input the artifact and outputs its final representation. This final representation could be a graphical representation or any other format as needed: image, sound, video, string, etc. In our maze generation example, the function will return a black-and-white image where white pixels are passable tiles and black pixels are solid tiles.
\end{itemize}
Besides these functions, the content space and control space have to be defined, since each problem can have its own complex representation and control parameter. For example, the \emph{Arcade Rules} problem (see Section \ref{sec:protocol_problems}) has a representation that explains the rules of the game and game object locations and a control parameter which is a 2D layout of solid and empty tiles which constitutes the level that the rules should work on.

\section{Experiments}\label{sec:protocol}
This section covers the experimental protocol used to test our benchmark\footnote{Available at \url{https://github.com/amidos2006/benchmark_experiments}}. In Section \ref{sec:protocol_problems}, we describe each of the generative problems presented in the benchmark. Section~\ref{sec:protocol_generators} focuses on describing the baseline generators used to test them. Finally, Section~\ref{sec:protocol_fitness} explains the representation used and the different fitness functions used during our experiments.

\subsection{Problems} \label{sec:protocol_problems}
The PCG Benchmark includes 12 PCG problems, each with its own representation, control parameters, quality, diversity, and controllability criteria. We will summarize all of these problems here but more details are found in the documentation\footnote{Available at \url{https://github.com/amidos2006/pcg_benchmark}}. 
 
\textbf{Arcade Rules} hinges on generating the rules for new 2D arcade games based on the framework provided in \cite{togelius_2008_experiment}. The problem consists of a dictionary of integers (including the starting coordinates, winning conditions, movement rules, and collision rules) mapped to different effects. For example, if the win condition is $0$, the game is won if the player is alive after $40$ frames since the game starts. Seven quality criteria ensure that the game can be won, lost, and is playable by several different agents (e.g. static or random) that reach different performance profiles. The control parameter is a $7\times7$ binary array of the level layout (each tile is solid or empty); if not provided, a simple fixed layout is used.

\textbf{Binary} originates from the PCGRL framework \cite{khalifa_2020_pcgrl}, and involves generating fully connected 2D mazes, consisting of empty or solid tiles. The default variant of the problem is to generate $14\times14$ mazes, and the quality constraint is having a minimum length of the longest path between any two empty tiles in the maze of at least $28$ tiles. The problem has one control parameter, further specifying the minimum length of the longest path.

\textbf{Building} is inspired by \cite{jiang_2022_learning}, which presents the challenge of generating buildings in a 3D space with four different types of Lego blocks ($1\times1$, $1\times3$, $3\times1$, and $3\times3$ voxels). The default variant of the problem cares about generating buildings that use $40$ Lego block types, must have a maximum size of $7\times7\times12$ voxels, and must be taller than $6$ voxels (quality constraints). Four control parameters determine the ratio of different blocks to be used (out of $40$).

\textbf{Dangerous Dave} hinges on generating level layouts for a small discrete version of the DOS game \emph{Dangerous Dave} (Uptime Disk Monthly, 1988). The game is a small platformer where the player must avoid spikes, collect diamonds, and reach the exit. Generated levels have multiple quality constraints (number of tiles per type, solvability by AI agents, minimum number of jumps, and reachability of all diamonds). The default variant of the problem has $11\times7$ tiles and the solution must have at least $2$ jumps. The problem has five control parameters (coordinates of start tile and exit tile, and number of diamonds required).

\textbf{Elimination} uses the word game \emph{Elimination} (Khalifa, 2018) described in \cite{khalifa_2019_elimination}. The problem is to generate a sequence of letters that can create at least one short word, one long word, and nothing longer. The problem has five quality constraints (including words allowed). The default variant of this problem presents sequences of 8 letters and requires the short words to lie between 40\% and 60\% of the most common English words and the long words to lie between 60\% to 80\% of the most common English words. The problem has one control parameter, determining the maximum consecutive letters of an actual word in the initial sequence.

\textbf{Isaac} hinges on generating fully connected dungeons for a simplified version of \emph{The Binding of Isaac} (McMillen, 2011) video game. The dungeons generated must contain a target number of rooms, including a starting room, a boss room, a treasure room, and a shop room. Five quality constraints assess whether the dungeon is connected, whether all special rooms are present, the minimum distance between certain rooms, etc. The problem has one control parameter (the number of rooms in the dungeon).

\textbf{Lode Runner} is based on a simplified version of the \emph{Lode Runner} (Broderbund, 1983) puzzle platformer video game. The goal is to generate a playable level with a minimum number of gold and enemies. Generated levels must fulfill seven quality constraints, including minimum tiles per type, and having most tiles reachable to the player. The problem has two control parameters (the number of ladder tiles and rope tiles). The default variant of this problem requires levels of size $32\times22$ tiles, encoded as a grid of $16\times11$ indices (with $2\times2$ tiles per index), with a minimum of $6$ gold items and $3$ enemies.

\textbf{MiniDungeons} is based on the \emph{MiniDungeons} framework \cite{liapis_2015_procedural}, where the aim of the game is to reach the exit without dying to enemies. This variant uses deterministic combat when facing enemies. Four quality constraints define the minimum number of tiles per type in the level, whether the dungeon is connected, and whether at least some monsters are defeated along the shortest path to the exit. The default variant of the problem is to generate a solvable $8\times12$ level which forces the player to kill $12$ enemies before reaching the exit. The problem has five control parameters (coordinates of start tile and exit tile, target number of treasures).
    
\textbf{Super Mario Bros} is inspired by the work done in \cite{dahlskog_2014_linear} where \emph{Super Mario Bros.} levels are represented as a sequence of vertical slices sampled from the original game. The default variant of the problem is to generate a playable level made of $150$ slices with a similar look to original Mario levels. Five quality constraints assess whether the level can be completed by an A* agent, has flat areas or small elevations, unbroken pipe structures, and few floating enemies. The problem has three control parameters (number of enemies, coins, and jumps achieved during an AI playthrough).

\textbf{Sokoban} is based on the Japanese block-pushing game (Imabayashi, 1982) and a generator must create fully solvable puzzles. The default variant of this problem requires generating playable levels of $5\times5$ tiles; five quality constraints test the presence of special tiles (player, crate) and that an A* agent can solve the level in at least $10$ moves. The problem has one control parameter (target number of crates).

\textbf{Talakat} generates bullet patterns for the Talakat shoot-em-up game \cite{khalifa_2018_talakat}. Four quality constraints assess the quality of the bullet patterns (e.g. maximum bullet spawners, minimum bullets per frame, distribution of bullets in different parts of the screen, etc.). The problem's control parameter is a 1D array of the distribution of bullets over time within one second of gameplay.

\begin{figure*}[t]
    \centering
    \includegraphics[width=0.88\linewidth]{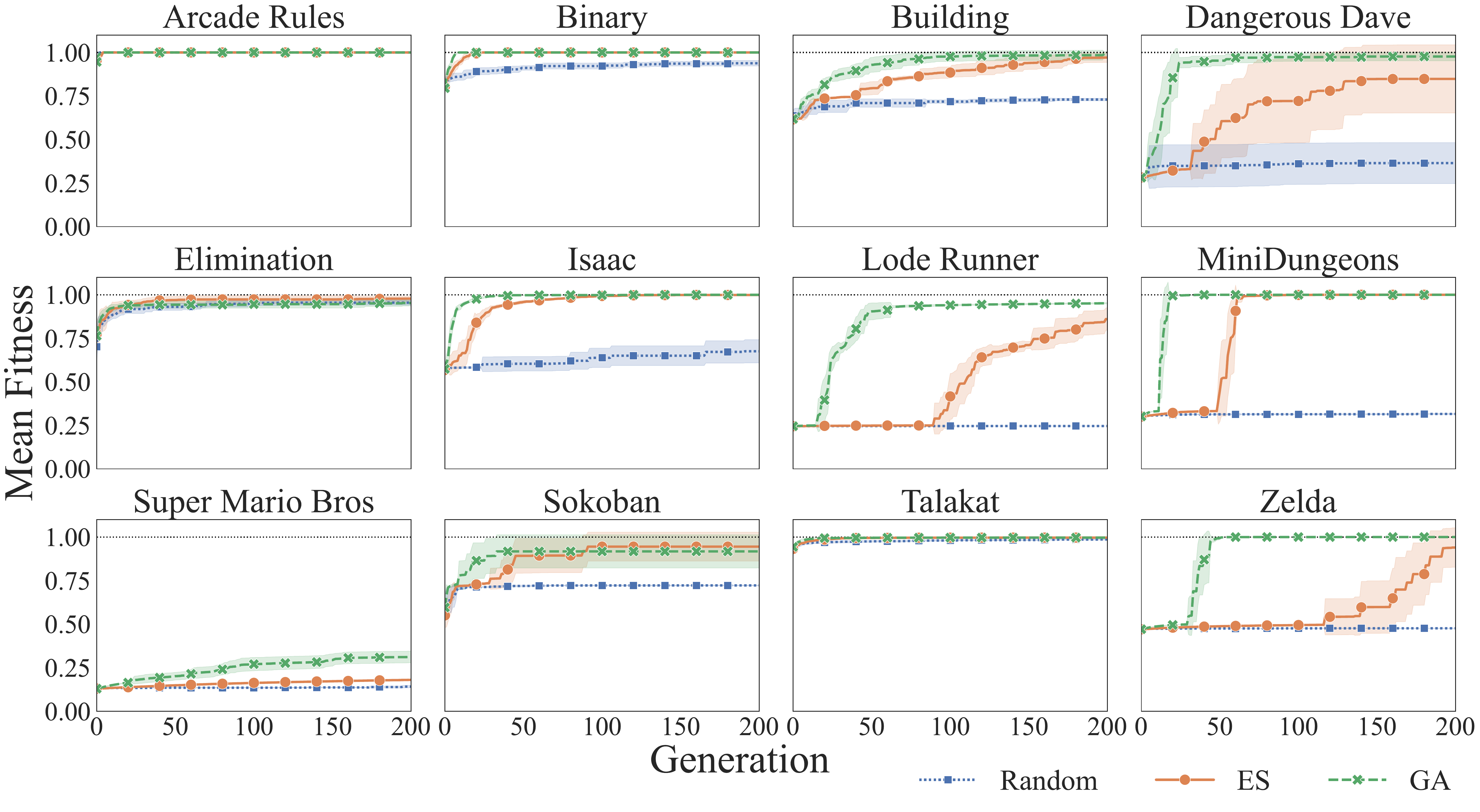}
    \caption{Progression of the maximum fitness when optimizing the Quality fitness with the three baseline algorithms. Results are averaged from 10 runs, with 95\% confidence intervals as the shaded area.}
    \label{fig:quality_progression}
\end{figure*}

\textbf{Zelda} originates from the General Video Gameplaying AI framework \cite{perez_2019_general}, and has been used in several research papers \cite{khalifa_2020_pcgrl, torrado_2020_bootstrapping, siper_2022_path}. The generated level is a maze with a key, a door, and enemies: the player must find a key to get to the exit without dying. Six quality constraints test the connectivity of the level, the number of tiles of each type, and the minimum length for the solution. The problem has two control parameters (the minimum distance from the start to the key tile and the minimum distance from the key to the door tile).

\subsection{Generators} \label{sec:protocol_generators}
In this initial study, we test three baseline generators on the PCG Benchmark. All generators follow simple \emph{search-based} PCG approaches \cite{togelius_2011_search}. Search-based algorithms can be used for black-box optimization; they do not need to understand (or adapt to) the problem~\cite{audet2022blackbox}. The generators were tested on the default variants for all of the described problems, and each experiment was run in 10 independent runs. All methods were required to produce 100 individuals during each generation. All the generators ran for 200 generations. The implementation of each generator is as follows:
\begin{itemize}
\item \textbf{Random Generator (Random):} Every generation, a new population of individuals is randomly created and evaluated. The new population is combined with the previous generation, and the best 100 individuals are kept for this generation.
\item \textbf{$\mu + \lambda$ Evolutionary Strategy (ES):} Every generation, a new population of 100 ($\lambda$) is mutated from the previous generation and evaluated, with the best 100 ($\mu$) individuals kept between generations. Candidates are evolved using a uniform mutation rate of 5\%. 
\item \textbf{Genetic Algorithm (GA):} This method extends ES by introducing selection and crossover operators. We use tournament selection on 7 individuals to select candidates that produce new offspring. Uniform crossover is used to combine parents, with a crossover rate of 50\%. Offspring have a 5\% mutation rate, same as ES. The population size is 100 and elitism preserves the best 10 solutions between generations.
\end{itemize}

\subsection{Representation and Fitness Functions}\label{sec:protocol_fitness}
The chromosome for all three algorithms consists of two parts, a content vector and a control parameter vector. These are sampled at the beginning randomly from the content space and control space (explained in Section~\ref{sec:pcg_benchmark}). For the sake of simplicity, operators only work on the content part; nothing changes the control parameter (however, different solutions satisfy different controllability constraints). Looking into the fitness function, we use 3 different fitness functions that care about quality, controllability, and diversity:
\begin{itemize}
\item \textbf{Quality Fitness (Q):} Solution fitness is equal to the quality of the artifact. This fitness does not check if controllability constraints are met. The formula is shown in Eq.~\eqref{eq:q_fitness}. 
\begin{equation}\label{eq:q_fitness}
f(c_i,p_i,\mathbb{C}) = q(c_i)
\end{equation}
\noindent where $c_i$ is the content being evaluated, $p_i$ is the control parameter associated, $\mathbb{C}$ is the total population of content, and $q(c_i)$ is the quality value for input content.
\item \textbf{Quality then Controllability Fitness (QT):} This fitness function first tries to maximize quality, then controllability if the solution is optimal with regard to quality. The fitness is shown in Eq.~\eqref{eq:qc_fitness}.
\begin{equation}\label{eq:qc_fitness}
f(c_i,p_i, C) = \begin{cases}
    \frac{1}{2}q(c_i) & \text{if $q(c_i) < 1$}\\
    \frac{1}{2}(q(c_i) + t(c_i, p_i)) & \text{if $q(c_i) = 1$}
\end{cases}
\end{equation}
\noindent where $c_i$ is the content being evaluated, $p_i$ is the control parameter associated, $C$ is the total population of content, $q(c_i)$ is the quality value for input content, and $t(c_i, p_i)$ is the controllability value for the input content with respect to the input control parameter.
\item \textbf{Quality, Controllability, then Population Diversity Fitness (QTD):} Similar to the previous function, this fitness function first tries to maximize quality, then after that controllability if the solution is optimal with regard to quality. Finally, if the solution is optimal with respect to quality and controllability, it tries to optimize towards population diversity as shown in Eq.~\eqref{eq:qcd_fitness}. Population diversity is challenging as it fluctuates depending on the current population, but it can help search algorithms overcome local optima.
\begin{equation}\label{eq:qcd_fitness}
f(c_i, p_i, C) = \begin{cases}
    \frac{1}{3}q(c_i) & 
\begin{array}{rl}
\text{if~} &q(c_i) < 1,\\ 
&t(c_i,p_i) < 1\\
\end{array}
    \\
    \frac{1}{3}(q(c_i) + t(c_i, p_i)) & 
\begin{array}{rl}
\text{if~} &q(c_i) = 1,\\ 
&t(c_i,p_i) < 1\\
\end{array}
    \\
    \frac{1}{3}\left(q(c_i) + t(c_i, p_i) + d(c_i, \mathbb{C})\right) & 
\begin{array}{rl}
\text{if~} &q(c_i) = 1,\\ 
&t(c_i,p_i) = 1\\
\end{array}
\end{cases}
\end{equation}
\noindent where $c_i$ is the content being evaluated, $p_i$ is the control parameter associated, $C$ is the total population of content, $q(c_i)$ is the quality value for input content, $t(c_i, p_i)$ is the controllability value for the input content with respect to the input control parameter, and $d(c_i, \mathbb{C})$ is a measure of uniqueness of $c_i$ with respect to the set of content $\mathbb{C}$ in the population.
\end{itemize}

\begin{figure*}[t]
    \centering
    \includegraphics[width=0.88\linewidth]{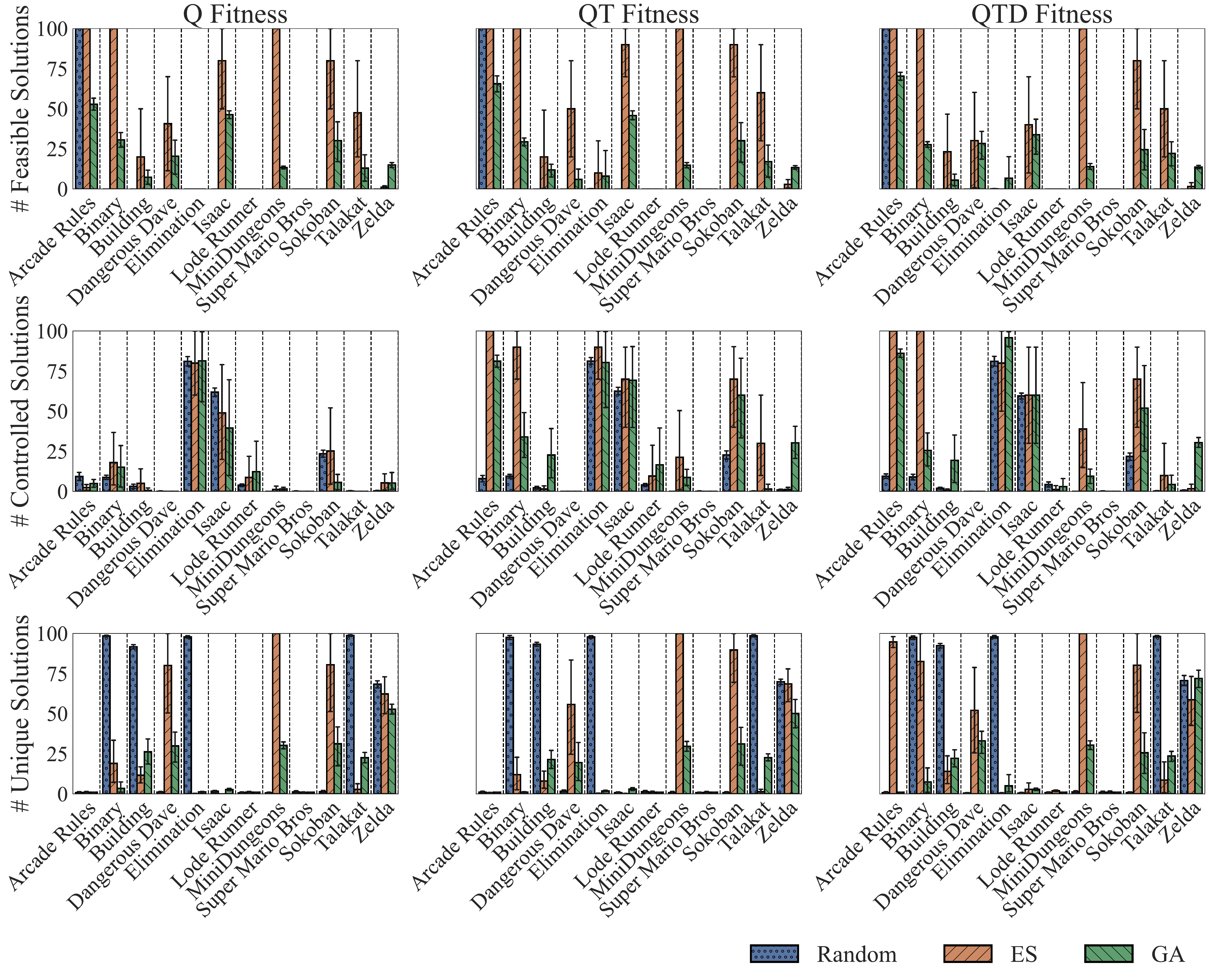}
    \caption{Number of solutions $c_i$ (out of 100) in the final population (after 200 generations) that are feasible ($q(c_i) = 1$), controlled ($t(c_i, p_i)=1$), and unique ($d(c_i, \mathbb{C})=1$ compared to the final population $\mathbb{C}$). Results are averaged from 10 runs, with 95\% confidence intervals as error bars.}
    \label{fig:success_graphs}
\end{figure*}

\section{Results}

In our experiment, we test three search-based algorithms (see Section \ref{sec:protocol_generators}) using three different fitness functions to optimize content for different problems. Each problem has its own measures for what constitutes good (i.e. \emph{feasible} content $c_i$ if $q(c_i) = 1$) and diverse (i.e. \emph{unique} content if its $d(c_i, \mathbb{C})=1$ with respect to a set of content $\mathbb{C}$). Moreover, in the initial population, each individual has its own control parameters randomized within allowed value ranges. While the control parameters are not randomized, mutated, or recombined \emph{during} evolution, the search process will prioritize copying over (via elitism, $\mu$+$\lambda$, or choosing the best 100 individuals in Random) solutions with more easily satisfied control parameters. Each problem similarly has its own controllability score, based on the content $c_i$ and its paired control parameters $p_i$: controllability criteria are satisfied if $t(c_i, p_i)=1$; as shorthand, we label such individuals \emph{controlled} in the below section. 

Figure~\ref{fig:quality_progression} shows how the maximum fitness (when optimizing for Quality alone) improves over the generations for the different algorithms. We observe that some problems can discover feasible solutions even with random initialization: \emph{Arcade Rules} and \emph{Talakat} have near-optimal individuals in the initial population, which leads to almost no improvement from evolution. Some problems struggle to discover high-quality individuals with random initialization: \textit{Lode Runner}, \emph{MiniDungeons} and \emph{Dangerous Dave} start with a maximum fitness around 0.25 before evolution, and \emph{Super Mario Bros} start from around 0.13. Evolution improves the maximum fitness in all problems, although to different degrees depending on the problem. GA has overall slightly better improvements in maximum fitness from initial to final population (between 6\% relative increase in maximum fitness for \emph{Talakat} and 288\% relative increase for \emph{Lode Runner}), with ES a close second (between 5\% relative increase in \emph{Arcade Rules} and 251\% relative increase for \emph{Lode Runner}). These relative increases can be noticed in the generated examples in tables \ref{tab:examples_quality_1} and \ref{tab:examples_quality_2} where the best random content looks subjectively more noisy compared to its counterpart (especially in \textit{Lode Runner}, \textit{Super Mario}, and \textit{Zelda}). Random does not increase maximum fitness as much, with the highest relative increase of 36\% from the initial population in \emph{Elimination} and negligible increases (below 1\%) in \emph{Lode Runner} and \emph{Zelda}. In \emph{Super Mario Bros}, however, no solution after 200 generations satisfies the quality constraints for any of the generators in any of the runs. In this problem, changes in the maximum fitness are overall slow (141\%, 39\%, and 9\% relative increase from the initial population for GA, ES, and Random respectively). This indicates that either the variation operators implemented or the quality constraints pose challenges to any stochastic search approach. Given similar difficulties in generating other large platformer levels in \emph{Lode Runner}, the size of the level (genotype) for \emph{Super Mario Bros} could also be an issue.

While so far we focused on how easy it is to generate feasible content with random initialization and to improve on them via search-based PCG, we are ultimately most interested in the artifacts at the end of the process (after 200 generations). Figure \ref{fig:success_graphs} shows the number of feasible individuals ($q(c_i)=1$) in the final population, the number of unique individuals ($d(c_i, \mathbb{C})=1$ where $\mathbb{C}$ is the final population in the same run), and the number of final individuals that satisfy all controllability constraints for their paired control parameters ($p_i$), i.e. $t(c_i, p_i)=1$.

\begin{table*}
    \centering
    \resizebox{\linewidth}{!}{%
        \begin{tabular}{m{.08\linewidth}>{\raggedright\arraybackslash}m{.15\linewidth}>{\raggedright\arraybackslash}m{.15\linewidth}|>{\raggedright\arraybackslash}m{.15\linewidth}>{\raggedright\arraybackslash}m{.15\linewidth}|>{\raggedright\arraybackslash}m{.15\linewidth}>{\raggedright\arraybackslash}m{.15\linewidth}}
            & \multicolumn{2}{c|}{Random Search} & \multicolumn{2}{c|}{$\mu + \lambda$ Evolution Strategy} & \multicolumn{2}{c}{Genetic Algorithm}\\
            Arcade Rules & \includegraphics[width=\linewidth]{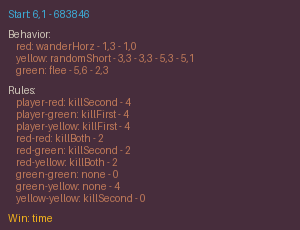} & \includegraphics[width=\linewidth]{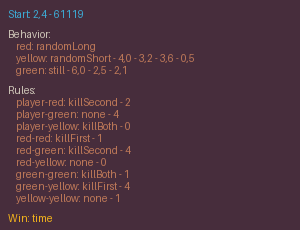} &
            \includegraphics[width=\linewidth]{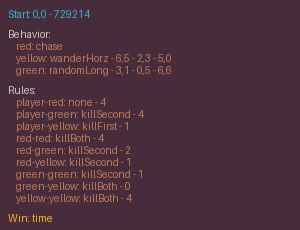} & \includegraphics[width=\linewidth]{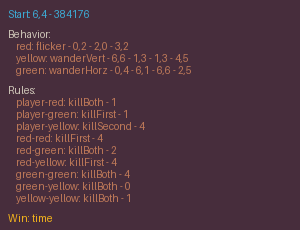} & 
            \includegraphics[width=\linewidth]{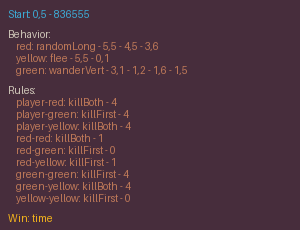} & \includegraphics[width=\linewidth]{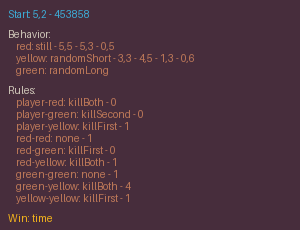}\\ 
            \multirow{1}*{Binary} & \cellcolor{red!25}\includegraphics[width=\linewidth]{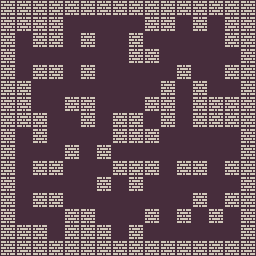} & \cellcolor{red!25}\includegraphics[width=\linewidth]{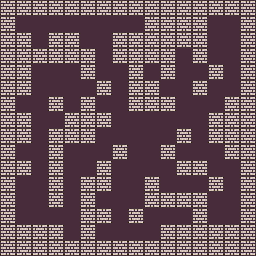} & 
            \includegraphics[width=\linewidth]{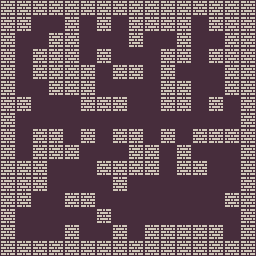} & \includegraphics[width=\linewidth]{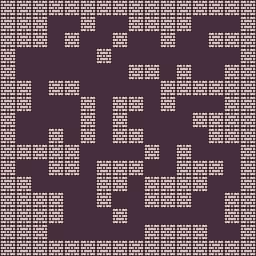} & 
            \includegraphics[width=\linewidth]{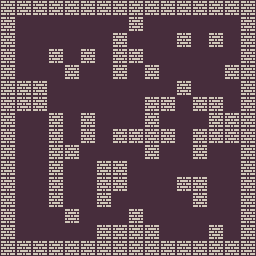} & \includegraphics[width=\linewidth]{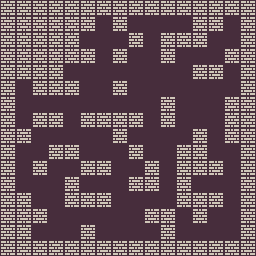} \\ 
            Building & \cellcolor{red!25}\includegraphics[width=\linewidth]{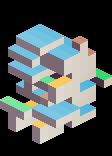} & \cellcolor{red!25}\includegraphics[width=\linewidth]{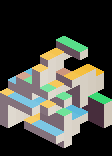} & 
            \includegraphics[width=\linewidth]{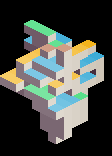} & \includegraphics[width=\linewidth]{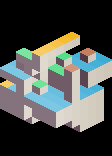} & 
            \includegraphics[width=\linewidth]{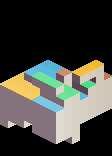} & \includegraphics[width=\linewidth]{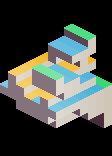}\\
            Dangerous Dave & \cellcolor{red!25}\includegraphics[width=\linewidth]{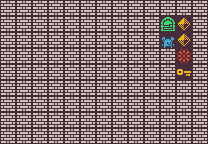} & \cellcolor{red!25}\includegraphics[width=\linewidth]{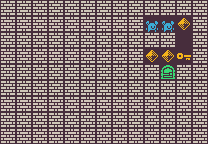} & 
            \includegraphics[width=\linewidth]{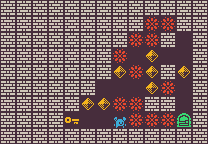} & \includegraphics[width=\linewidth]{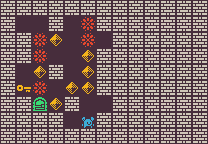} & 
            \includegraphics[width=\linewidth]{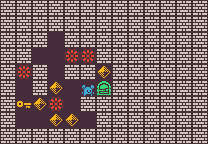} & \includegraphics[width=\linewidth]{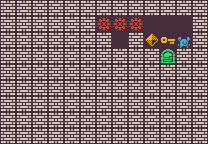}\\
            Elimination & \cellcolor{red!25}\includegraphics[width=\linewidth]{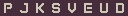} & \cellcolor{red!25}\includegraphics[width=\linewidth]{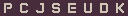} & 
            \cellcolor{red!25}\includegraphics[width=\linewidth]{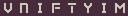} & \cellcolor{red!25}\includegraphics[width=\linewidth]{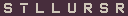} & 
            \cellcolor{red!25}\includegraphics[width=\linewidth]{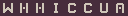} & \cellcolor{red!25}\includegraphics[width=\linewidth]{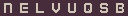}\\ 
            Isaac & \cellcolor{red!25}\includegraphics[width=\linewidth]{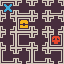} & \cellcolor{red!25}\includegraphics[width=\linewidth]{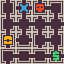} & 
            \includegraphics[width=\linewidth]{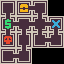} & \includegraphics[width=\linewidth]{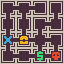} & 
            \includegraphics[width=\linewidth]{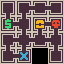} & \includegraphics[width=\linewidth]{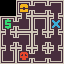}\\
        \end{tabular}%
    }
    \caption{Example of the best generated content (one per run) for 3 different generators optimizing the \textit{Quality} fitness function for the first 6 problems (see Section \ref{sec:protocol_problems}). Content with a red background means it failed the quality constraints.}
    \label{tab:examples_quality_1}
\end{table*}

\begin{table*}
    \centering
    \resizebox{\linewidth}{!}{%
        \begin{tabular}{m{.08\linewidth}>{\raggedright\arraybackslash}m{.15\linewidth}>{\raggedright\arraybackslash}m{.15\linewidth}|>{\raggedright\arraybackslash}m{.15\linewidth}>{\raggedright\arraybackslash}m{.15\linewidth}|>{\raggedright\arraybackslash}m{.15\linewidth}>{\raggedright\arraybackslash}m{.15\linewidth}}
            & \multicolumn{2}{c|}{Random Search} & \multicolumn{2}{c|}{$\mu + \lambda$ Evolution Strategy} & \multicolumn{2}{c}{Genetic Algorithm}\\
           \makecell[l]{Lode\\Runner} & \cellcolor{red!25}\includegraphics[width=\linewidth]{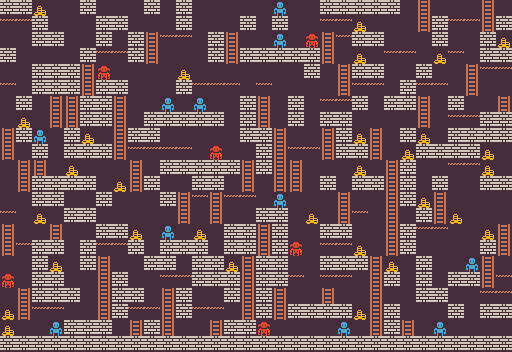} & \cellcolor{red!25}\includegraphics[width=\linewidth]{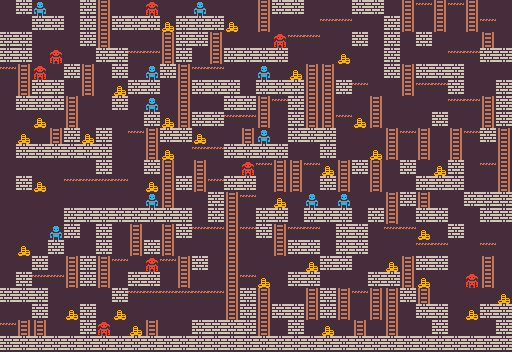} & 
            \cellcolor{red!25}\includegraphics[width=\linewidth]{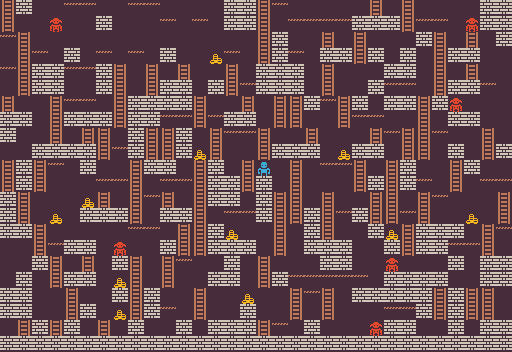} & \cellcolor{red!25}\includegraphics[width=\linewidth]{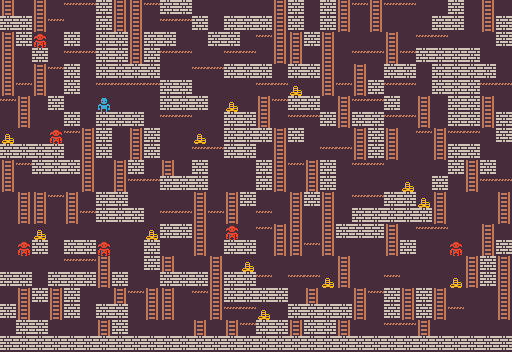} & 
            \cellcolor{red!25}\includegraphics[width=\linewidth]{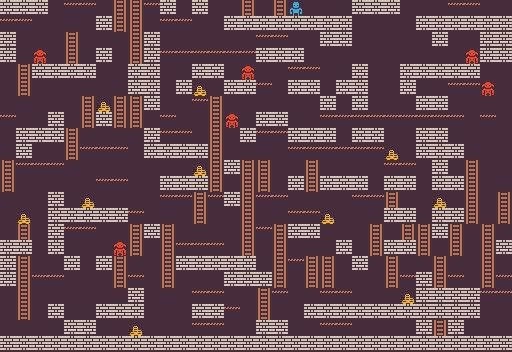} & \cellcolor{red!25}\includegraphics[width=\linewidth]{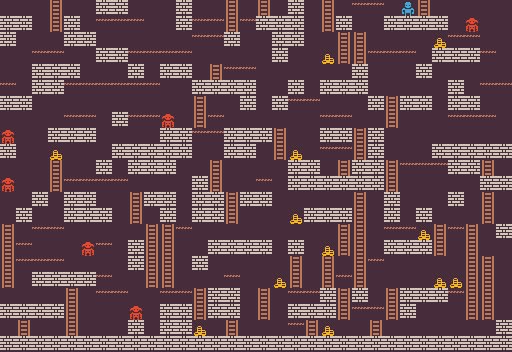}\\
            \makecell[l]{Mini\\Dungeons} & \cellcolor{red!25}\includegraphics[width=\linewidth]{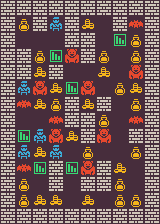} & \cellcolor{red!25}\includegraphics[width=\linewidth]{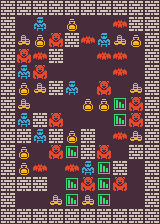} & 
            \includegraphics[width=\linewidth]{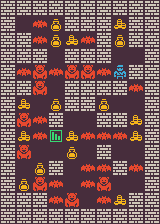} & \includegraphics[width=\linewidth]{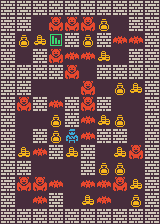} & 
            \includegraphics[width=\linewidth]{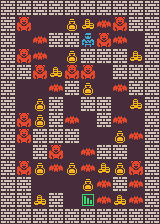} & \includegraphics[width=\linewidth]{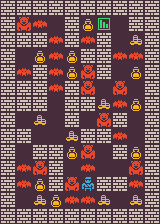}\\
            Super Mario Bros & \cellcolor{red!25}\includegraphics[width=\linewidth]{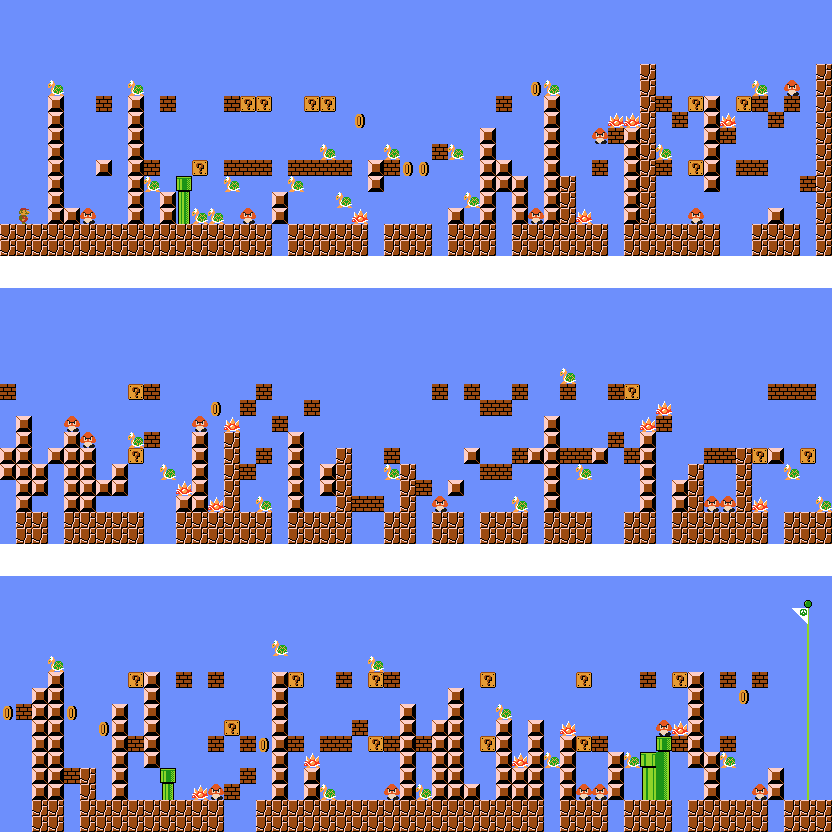} & \cellcolor{red!25}\includegraphics[width=\linewidth]{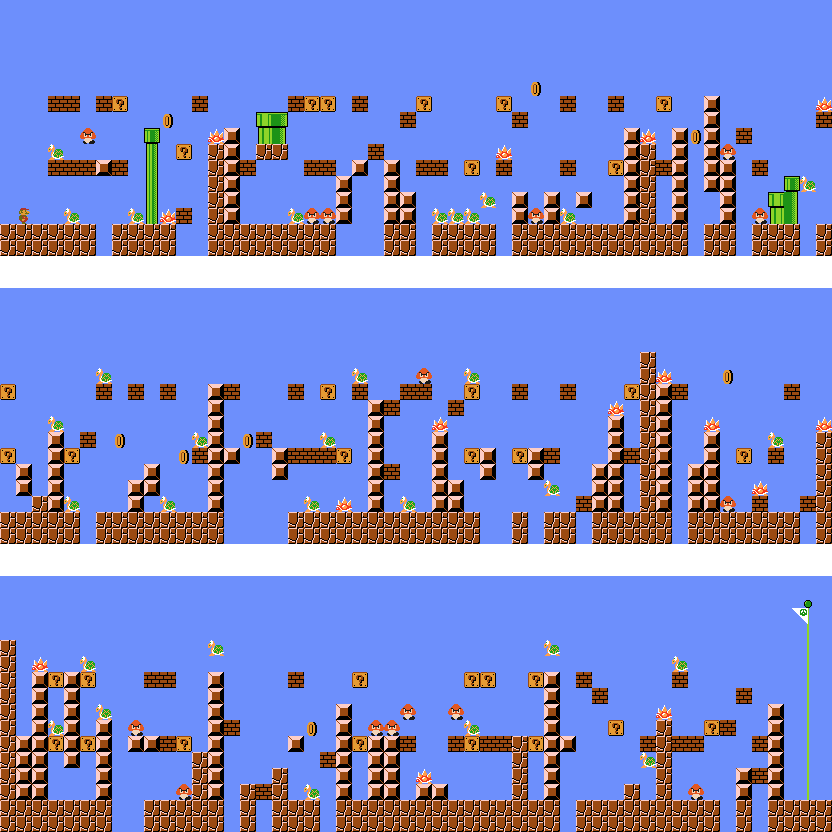} & 
            \cellcolor{red!25}\includegraphics[width=\linewidth]{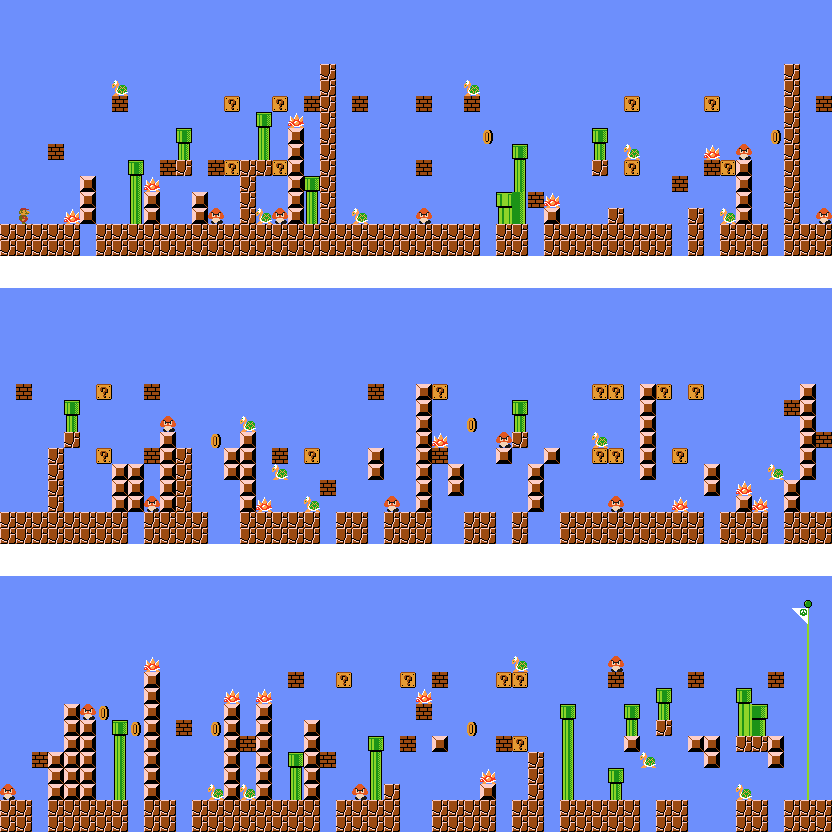} & \cellcolor{red!25}\includegraphics[width=\linewidth]{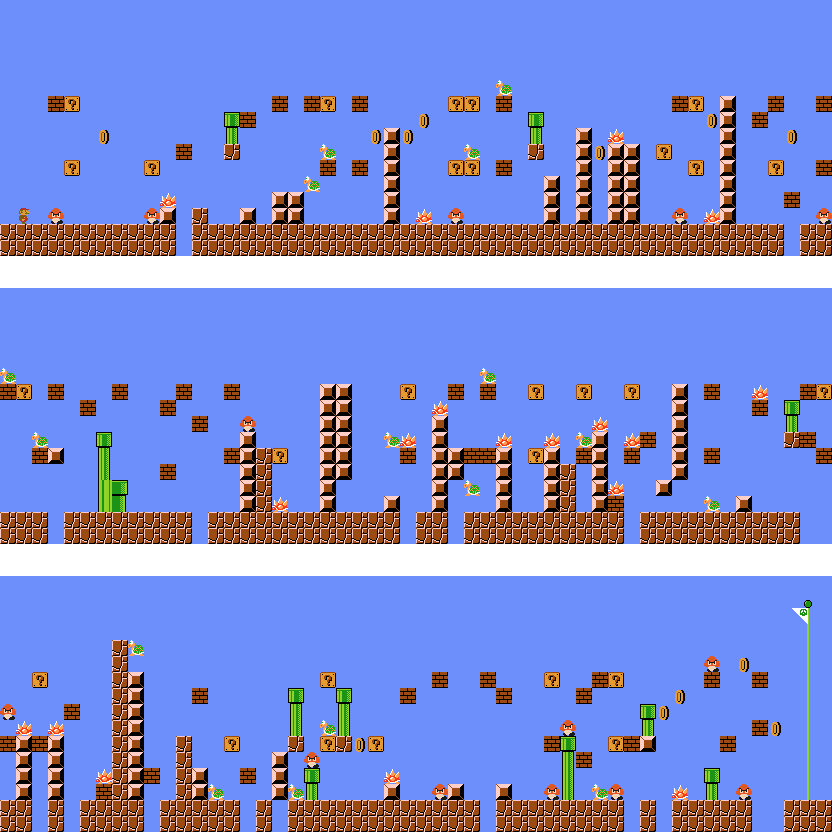} & 
            \cellcolor{red!25}\includegraphics[width=\linewidth]{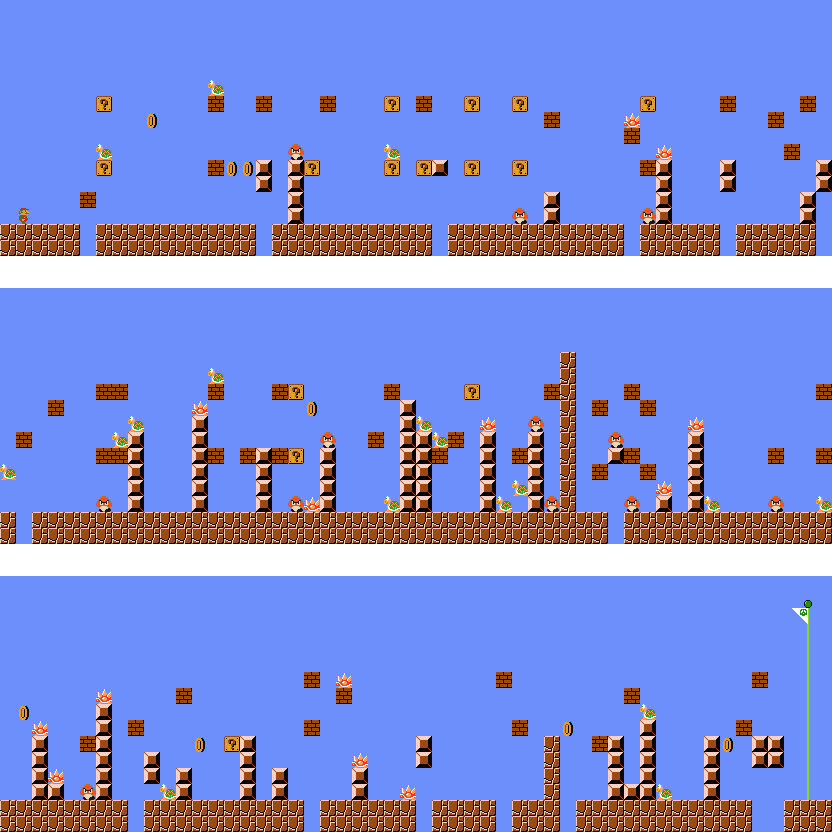} & \cellcolor{red!25}\includegraphics[width=\linewidth]{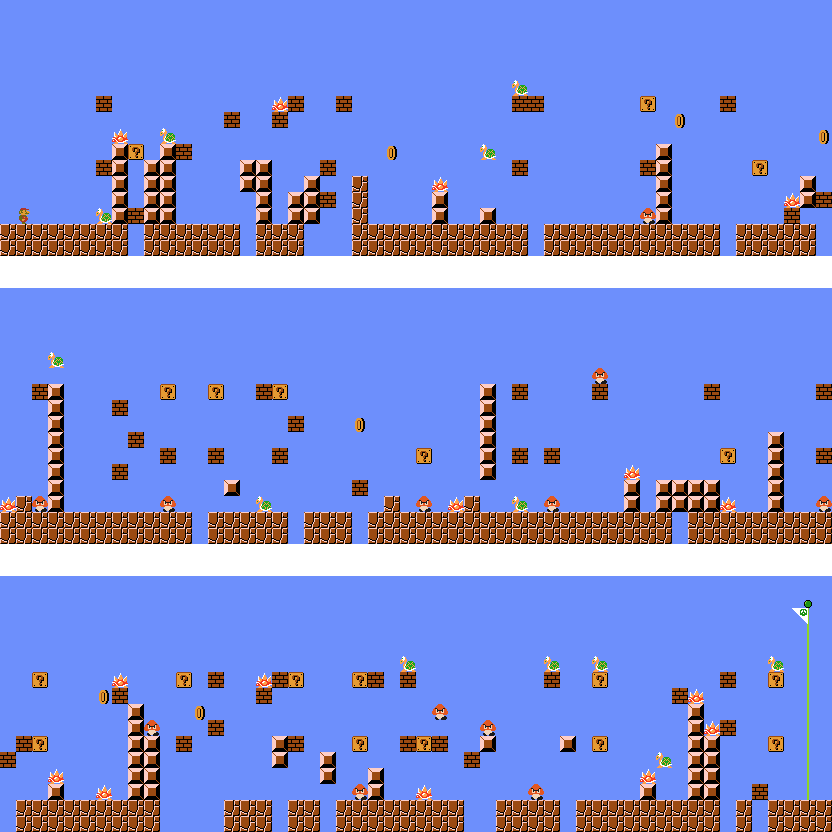}\\
            Sokoban & \cellcolor{red!25}\includegraphics[width=\linewidth]{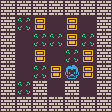} & \cellcolor{red!25}\includegraphics[width=\linewidth]{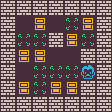} & 
            \includegraphics[width=\linewidth]{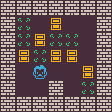} & \includegraphics[width=\linewidth]{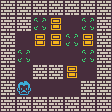} & 
            \includegraphics[width=\linewidth]{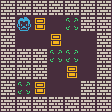} & \includegraphics[width=\linewidth]{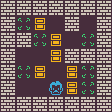}\\
            Talakat & \cellcolor{red!25}\includegraphics[width=\linewidth]{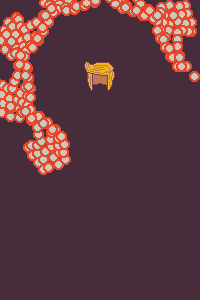} & \cellcolor{red!25}\includegraphics[width=\linewidth]{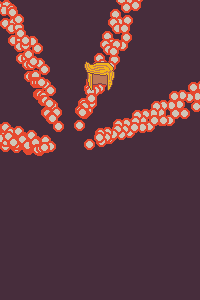} &
            \includegraphics[width=\linewidth]{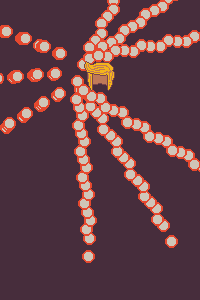} & \includegraphics[width=\linewidth]{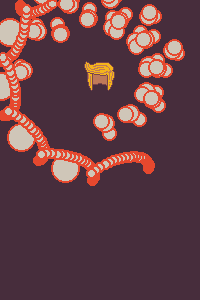} &
            \includegraphics[width=\linewidth]{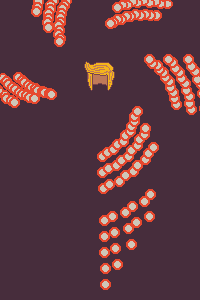} & \includegraphics[width=\linewidth]{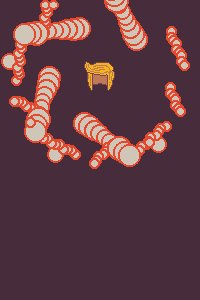}\\
            Zelda & \cellcolor{red!25}\includegraphics[width=\linewidth]{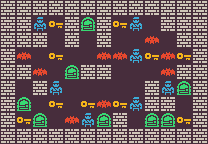} & \cellcolor{red!25}\includegraphics[width=\linewidth]{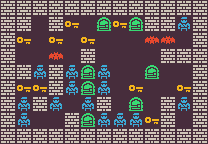} & 
            \includegraphics[width=\linewidth]{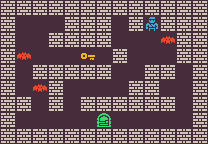} & \includegraphics[width=\linewidth]{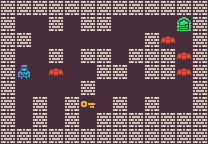} &
            \includegraphics[width=\linewidth]{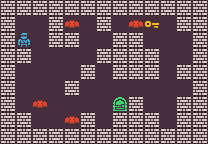} & \includegraphics[width=\linewidth]{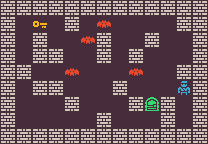}\\
        \end{tabular}%
    }
    \caption{Example of the best generated content (one per run) for 3 different generators optimizing the \textit{Quality} fitness function for the last 6 problems (see Section \ref{sec:protocol_problems}). Content with a red background means it failed the quality constraints.}
    \label{tab:examples_quality_2}
\end{table*}

Observing the number of individuals that satisfy all quality constraints in the top row of Fig.~\ref{fig:success_graphs}, conclusions mirror those from the maximum Q fitness progression (Fig.~\ref{fig:quality_progression}). Some games are easy to find feasible individuals for: \emph{Arcade Rules}, \emph{Binary}, \emph{MiniDungeons} end up with an entire population of feasible individuals in all runs of ES regardless of fitness being optimized. Interestingly, in \emph{Arcade Rules} Random also reaches a final population full of feasible individuals, but no feasible individuals in the other three games. Other games are too hard to find feasible individuals for any algorithm or fitness target: \emph{Super Mario Bros} and \emph{Lode Runner} consistently have no feasible individuals in the final population. Interestingly, the choice of fitness target also impacts the number of feasible individuals: In \textit{Isaac} (and in \textit{Dangerous Dave} with ES) the number of feasible individuals dropped for QTD with respect to other fitness functions. Since the drop is not that big, it might be due to randomness. A special case is \emph{Elimination}, where optimizing Quality alone leads to no feasible individuals while optimizing either QT or QTD leads to feasible individuals (for ES and GA with QT, and for GA with QTD). Given that both QT and QTD fitnesses would have values of 0 for infeasible individuals, this difference in behavior is an artifact of randomization (either in initialization or stochastic search process): indeed, feasible individuals were found only in one run of each algorithm and in one runs of GA for QTD fitness. Although \textit{Zelda} has a small number of feasible solutions in the final population, the GA manages to find at least one feasible chromosome in 8 of 10 runs while ES does so in 8 runs for Q fitness, 6 runs for QT fitness and 5 runs for QTD fitness. GA in general does not contain many feasible individuals in the final population but is much more consistent in discovering feasible individuals than ES. Of a total of 360 runs across all environments and all fitness targets, the GA has feasible solutions in the final population in 224 runs, versus 191 runs for ES and 32 runs for Random. The fewer feasible solutions per population in the GA are likely due to the different elitism mechanisms: ES saves the best 100 solutions while GA saves the best 10 solutions.

Observing the number of individuals that satisfy all (paired) controllability constraints in the second row of Fig.~\ref{fig:success_graphs}, it is evident that some benchmarks are very difficult to satisfy controllability criteria. \emph{Super Mario Bros} and \emph{Dangerous Dave} rarely have controlled individuals for any algorithm and for any fitness target. \emph{Lode Runner} is interesting because a few individuals are always controlled in every run of Random (regardless of fitness target), while GA and ES have few runs where any controlled individuals end up in the final population even with QT fitness (1 run for ES, 3 runs for GA) and with QTD fitness (1 run for ES, 2 runs for GA). In most problems, applying pressure for controllability (via QT or QTD fitness) does have an effect, with higher numbers of controllable individuals than for Quality alone. An interesting finding is that for some benchmarks the QTD fitness (which has more objectives and is thus more diluted) leads to many more controlled individuals than QT fitness alone, especially \emph{Talakat} for GA and \emph{MiniDungeons} for ES. Worth noting is that, at least with QT and QTD fitness targets, some benchmarks are easy to find controlled individuals for (particularly \emph{Arcade Rules}, \emph{Binary}, \emph{Elimination}, \emph{Isaac}, and \emph{Sokoban}).

From the number of unique individuals in the final population of each run (third row of Fig.~\ref{fig:success_graphs}), we observe that some benchmarks are easy to find unique individuals for (even if not explicitly targeting that as an objective via QTD fitness). \emph{Binary}, \emph{Building}, \emph{Elimination}, and \emph{Talakat} achieve many unique individuals with Random, fewer unique individuals with GA, and few if any with ES: this holds both for Q and QT fitness targets, which do not target diversity as an objective. For QTD, Random outperforms the other stochastic search methods in \emph{Building}, \emph{Elimination}, and \emph{Talakat}. However, QTD pushes GA to maintain at least one unique individual in all runs of all problems (even if for \emph{Arcade Rules}, \emph{Lode Runner} and \emph{Super Mario Bros} it's always only one unique individual). Finally, \emph{Super Mario Bros}, \emph{Isaac} and \emph{Lode Runner} struggle to maintain many unique individuals with any algorithm even when targeting QTD; however, it is important to note that \emph{Isaac} manages to maintain many feasible and controlled individuals with this fitness target.

The above experiments highlight how different the problems included in the PCG Benchmark are in terms of generating feasible, unique, or controlled individuals. Results also indicate that different generators have different trade-offs in different problems: maximum fitness improves faster with more pressure towards convergence from the GA (see Fig.~\ref{fig:quality_progression}) while introducing random individuals (rather than modifying the population) leads to more unique final solutions at the cost of feasibility and controllability. While the goal of this experiment was not necessarily to compare algorithms, the $\mu+\lambda$ ES seems more suitable as a general approach with more feasible solutions, more unique solutions, and more controlled solutions overall. It is worth noting however that the GA is more reliable: considering all three fitness targets, the GA produces feasible solutions after 200 generations in 224 of 360 runs (10 runs, 12 problems, 3 fitness targets), compared to 191 runs for ES and 32 runs for Random. The extensibility of the current problems, however, opens more possibilities for follow-up experiments: indicative modifications could be applied to some harder problems (e.g. \emph{Super Mario Bros} and \emph{Lode Runner}) towards fewer quality constraints or smaller genotypes, or to hard-code controllability parameters rather than pair them with the solution as done in this paper.

\section{Conclusion}
This paper introduced a new benchmark to test any generative algorithm. The PCG Benchmark provides an interface that is easy to use and extend, similar to the \textit{OpenAI Gym}~\cite{brockman_2016_openai}. The framework comes with 12 problems from the get-go that span across multiple different domains (rules, levels, structures, words, patterns, etc). We tested three different baseline algorithms (random, evolutionary strategy, and genetic algorithm) against all 12 problems, guided by different fitness functions that incorporate quality, controllability, and population diversity criteria. We noticed that generating large levels (i.e.~\textit{Super Mario Bros} and \textit{Lode Runner}) was challenging for all baseline algorithms. Complex landscapes as in the \textit{Elimination} problem were also not easy to solve due to low locality \cite{togelius_2011_search}. Interestingly, combining quality measures with controllability and population diversity sometimes helped the algorithm find better solutions faster.

Overall, we believe that the PCG Benchmark is a first step towards comparing generative algorithms. We expect that it will assist PCG research and education, in a similar that the \textit{OpenAI Gym}~\cite{brockman_2016_openai} pushed RL research forward. We also believe that the benchmark is a great learning tool for newcomers to the field, as well as students. We note that solving a problem in the PCG Benchmark does not mean that this generative problem is solved for good or that the generated levels can be used for any type of human player; it means that the generator is good for a specific hypothetical player that the evaluation functions were designed for. For example, solving the \textit{Super Mario Bros} problem with its default parameters (see Section \ref{sec:protocol_problems}) does not mean that you have the best generator for Mario levels. Instead, it means you have a generator that can create playable Mario levels with 15 non-floating enemies that follow the same tile distribution as the original Mario levels. Even playability for Mario is not human but proxied using A* algorithms~\cite{karakovskiy_2012_mario}. We also want to note that the benchmark does not try to solve generality in the PCG domain. This means that the framework is not designed to have one agent that can tackle all problems but is more similar to \textit{OpenAI Gym}~\cite{brockman_2016_openai} where each problem has its own representation and can be solved. We decided not to tackle the generality problem for the sake of simplicity, usability, and learnability; there are other frameworks that tackle this problem~\cite{khalifa2016general}. The problems in the PCG Benchmark act as milestones for researchers and students; once reached, users can expand the PCG Benchmark with new problems or harder variants of current problems (by adjusting their parameters in Section \ref{sec:protocol_problems}). In the end, having a standardized way to compare generative algorithms will shed light on the strengths and the drawbacks of PCG algorithms, pushing towards novel solutions as exhibited in gameplaying AI~\cite{ecoffet2021first} to address hard exploration problems within the Arcade Learning Environment~\cite{bellemare_2013_arcade}.

\begin{acks}
Thanks to Kenny for creating 1-Bit Pack\footnote{Available at \url{https://kenney.nl/assets/1-bit-pack}} which was used for most of the 12 problems in the benchmark. Even the ones that did not use it were inspired by the color palette used in that pack.

This project has received funding from the Malta Council for Science and Technology (MCST) through the SINO-MALTA Fund 2022, Project OPtiMaL.
\end{acks}

\bibliographystyle{ACM-Reference-Format}
\bibliography{biblo}


\begin{thebibliography}{60}


\ifx \showCODEN    \undefined \def \showCODEN     #1{\unskip}     \fi
\ifx \showISBNx    \undefined \def \showISBNx     #1{\unskip}     \fi
\ifx \showISBNxiii \undefined \def \showISBNxiii  #1{\unskip}     \fi
\ifx \showISSN     \undefined \def \showISSN      #1{\unskip}     \fi
\ifx \showLCCN     \undefined \def \showLCCN      #1{\unskip}     \fi
\ifx \shownote     \undefined \def \shownote      #1{#1}          \fi
\ifx \showarticletitle \undefined \def \showarticletitle #1{#1}   \fi
\ifx \showURL      \undefined \def \showURL       {\relax}        \fi
\providecommand\bibfield[2]{#2}
\providecommand\bibinfo[2]{#2}
\providecommand\natexlab[1]{#1}
\providecommand\showeprint[2][]{arXiv:#2}

\bibitem[Audet(2022)]%
        {audet2022blackbox}
\bibfield{author}{\bibinfo{person}{Charles Audet}.}
  \bibinfo{year}{2022}\natexlab{}.
\newblock \showarticletitle{Blackbox optimization}.
\newblock In \bibinfo{booktitle}{\emph{Encyclopedia of Optimization}}.
  \bibinfo{publisher}{Springer}.
\newblock


\bibitem[Barthet et~al\mbox{.}(2024)]%
        {barthet2024affectively}
\bibfield{author}{\bibinfo{person}{Matthew Barthet}, \bibinfo{person}{Roberto
  Gallotta}, \bibinfo{person}{Ahmed Khalifa}, \bibinfo{person}{Antonios
  Liapis}, {and} \bibinfo{person}{Georgios~N. Yannakakis}.}
  \bibinfo{year}{2024}\natexlab{}.
\newblock \showarticletitle{{Affectively Framework}: Towards Human-like
  Affect-Based Agents}. In \bibinfo{booktitle}{\emph{Proceedings of the
  International Conference on Affective Computing and Intelligent Interaction
  Workshops and Demos}}. \bibinfo{publisher}{IEEE}.
\newblock


\bibitem[Barthet et~al\mbox{.}(2022)]%
        {barthet_2022_open}
\bibfield{author}{\bibinfo{person}{Matthew Barthet}, \bibinfo{person}{Antonios
  Liapis}, {and} \bibinfo{person}{Georgios~N Yannakakis}.}
  \bibinfo{year}{2022}\natexlab{}.
\newblock \showarticletitle{Open-ended evolution for {Minecraft} building
  generation}.
\newblock \bibinfo{journal}{\emph{Transactions on Games}} \bibinfo{volume}{15},
  \bibinfo{number}{4} (\bibinfo{year}{2022}).
\newblock


\bibitem[Bellemare et~al\mbox{.}(2013)]%
        {bellemare_2013_arcade}
\bibfield{author}{\bibinfo{person}{Marc~G Bellemare}, \bibinfo{person}{Yavar
  Naddaf}, \bibinfo{person}{Joel Veness}, {and} \bibinfo{person}{Michael
  Bowling}.} \bibinfo{year}{2013}\natexlab{}.
\newblock \showarticletitle{The {Arcade Learning Environment}: An evaluation
  platform for general agents}.
\newblock \bibinfo{journal}{\emph{Journal of Artificial Intelligence Research}}
   \bibinfo{volume}{47} (\bibinfo{year}{2013}).
\newblock


\bibitem[Beukman et~al\mbox{.}(2022)]%
        {beukman_2022_procedural}
\bibfield{author}{\bibinfo{person}{Michael Beukman},
  \bibinfo{person}{Christopher~W Cleghorn}, {and} \bibinfo{person}{Steven
  James}.} \bibinfo{year}{2022}\natexlab{}.
\newblock \showarticletitle{Procedural content generation using neuroevolution
  and novelty search for diverse video game levels}. In
  \bibinfo{booktitle}{\emph{Proceedings of the Genetic and Evolutionary
  Computation Conference}}. \bibinfo{publisher}{ACM}.
\newblock


\bibitem[Bodria et~al\mbox{.}(2023)]%
        {bodria_2023_benchmarking}
\bibfield{author}{\bibinfo{person}{Francesco Bodria}, \bibinfo{person}{Fosca
  Giannotti}, \bibinfo{person}{Riccardo Guidotti}, \bibinfo{person}{Francesca
  Naretto}, \bibinfo{person}{Dino Pedreschi}, {and} \bibinfo{person}{Salvatore
  Rinzivillo}.} \bibinfo{year}{2023}\natexlab{}.
\newblock \showarticletitle{Benchmarking and survey of explanation methods for
  black box models}.
\newblock \bibinfo{journal}{\emph{Data Mining and Knowledge Discovery}}
  \bibinfo{volume}{37} (\bibinfo{year}{2023}).
\newblock


\bibitem[Brockman et~al\mbox{.}(2016)]%
        {brockman_2016_openai}
\bibfield{author}{\bibinfo{person}{Greg Brockman}, \bibinfo{person}{Vicki
  Cheung}, \bibinfo{person}{Ludwig Pettersson}, \bibinfo{person}{Jonas
  Schneider}, \bibinfo{person}{John Schulman}, \bibinfo{person}{Jie Tang},
  {and} \bibinfo{person}{Wojciech Zaremba}.} \bibinfo{year}{2016}\natexlab{}.
\newblock \showarticletitle{{OpenAI Gym}}.
\newblock \bibinfo{journal}{\emph{arXiv preprint arXiv:1606.01540}}
  \bibinfo{volume}{abs/1606.01540} (\bibinfo{year}{2016}).
\newblock


\bibitem[Buck(2015)]%
        {buck2015mazes}
\bibfield{author}{\bibinfo{person}{Jamis Buck}.}
  \bibinfo{year}{2015}\natexlab{}.
\newblock \bibinfo{booktitle}{\emph{Mazes for programmers: Code your own twisty
  little passages}}.
\newblock \bibinfo{publisher}{The Pragmatic Bookshelf}.
\newblock


\bibitem[Campbell et~al\mbox{.}(2002)]%
        {campbell_2002_deep}
\bibfield{author}{\bibinfo{person}{Murray Campbell}, \bibinfo{person}{A~Joseph
  Hoane~Jr}, {and} \bibinfo{person}{Feng-hsiung Hsu}.}
  \bibinfo{year}{2002}\natexlab{}.
\newblock \showarticletitle{Deep blue}.
\newblock \bibinfo{journal}{\emph{Artificial intelligence}}
  \bibinfo{volume}{134} (\bibinfo{year}{2002}).
\newblock


\bibitem[Charity et~al\mbox{.}(2020)]%
        {charity_2020_baba}
\bibfield{author}{\bibinfo{person}{Megan Charity}, \bibinfo{person}{Ahmed
  Khalifa}, {and} \bibinfo{person}{Julian Togelius}.}
  \bibinfo{year}{2020}\natexlab{}.
\newblock \showarticletitle{Baba is y’all: Collaborative mixed-initiative
  level design}. In \bibinfo{booktitle}{\emph{Proceedings of the Conference on
  Games}}. \bibinfo{publisher}{IEEE}.
\newblock


\bibitem[Compton and Mateas(2015)]%
        {compton2015casual}
\bibfield{author}{\bibinfo{person}{Kate Compton} {and} \bibinfo{person}{Michael
  Mateas}.} \bibinfo{year}{2015}\natexlab{}.
\newblock \showarticletitle{Casual Creators}. In
  \bibinfo{booktitle}{\emph{Proceedings of the International Conference on
  Computational Creativity}}. \bibinfo{publisher}{ICCC}.
\newblock


\bibitem[Cook(2019)]%
        {cook2019generative}
\bibfield{author}{\bibinfo{person}{Michael Cook}.}
  \bibinfo{year}{2019}\natexlab{}.
\newblock \bibinfo{title}{Tutorial: Generative \& Possibility Space}.
\newblock
  \bibinfo{howpublished}{\url{https://www.possibilityspace.org/tutorial-generative-possibility-space/index.html}}.
\newblock
\newblock
\shownote{Accessed On: 18-10-2024}.


\bibitem[Dahlskog et~al\mbox{.}(2014)]%
        {dahlskog_2014_linear}
\bibfield{author}{\bibinfo{person}{Steve Dahlskog}, \bibinfo{person}{Julian
  Togelius}, {and} \bibinfo{person}{Mark~J Nelson}.}
  \bibinfo{year}{2014}\natexlab{}.
\newblock \showarticletitle{Linear levels through n-grams}. In
  \bibinfo{booktitle}{\emph{Proceedings of the International Academic MindTrek
  Conference}}. \bibinfo{publisher}{ACM}.
\newblock


\bibitem[Ecoffet et~al\mbox{.}(2021)]%
        {ecoffet2021first}
\bibfield{author}{\bibinfo{person}{Adrien Ecoffet}, \bibinfo{person}{Joost
  Huizinga}, \bibinfo{person}{Joel Lehman}, \bibinfo{person}{Kenneth~O
  Stanley}, {and} \bibinfo{person}{Jeff Clune}.}
  \bibinfo{year}{2021}\natexlab{}.
\newblock \showarticletitle{First return, then explore}.
\newblock \bibinfo{journal}{\emph{Nature}}  \bibinfo{volume}{590}
  (\bibinfo{year}{2021}).
\newblock


\bibitem[Gallotta et~al\mbox{.}(2023)]%
        {gallotta2023preference}
\bibfield{author}{\bibinfo{person}{Roberto Gallotta}, \bibinfo{person}{Kai
  Arulkumaran}, {and} \bibinfo{person}{L.~B. Soros}.}
  \bibinfo{year}{2023}\natexlab{}.
\newblock \showarticletitle{Preference-Learning Emitters for Mixed-Initiative
  Quality-Diversity Algorithms}.
\newblock \bibinfo{journal}{\emph{Transactions on Games}}  \bibinfo{volume}{16}
  (\bibinfo{year}{2023}).
\newblock


\bibitem[Gallotta et~al\mbox{.}(2024)]%
        {gallotta2024large}
\bibfield{author}{\bibinfo{person}{Roberto Gallotta}, \bibinfo{person}{Graham
  Todd}, \bibinfo{person}{Marvin Zammit}, \bibinfo{person}{Sam Earle},
  \bibinfo{person}{Antonios Liapis}, \bibinfo{person}{Julian Togelius}, {and}
  \bibinfo{person}{Georgios~N. Yannakakis}.} \bibinfo{year}{2024}\natexlab{}.
\newblock \showarticletitle{Large Language Models and Games: A Survey and
  Roadmap}.
\newblock \bibinfo{journal}{\emph{IEEE Transactions on Games}}
  (\bibinfo{year}{2024}), \bibinfo{pages}{1--18}.
\newblock


\bibitem[Gravina et~al\mbox{.}(2019)]%
        {gravina_2019_procedural}
\bibfield{author}{\bibinfo{person}{Daniele Gravina}, \bibinfo{person}{Ahmed
  Khalifa}, \bibinfo{person}{Antonios Liapis}, \bibinfo{person}{Julian
  Togelius}, {and} \bibinfo{person}{Georgios~N Yannakakis}.}
  \bibinfo{year}{2019}\natexlab{}.
\newblock \showarticletitle{Procedural content generation through quality
  diversity}. In \bibinfo{booktitle}{\emph{Proceedings of the Conference on
  Games}}. \bibinfo{publisher}{IEEE}.
\newblock


\bibitem[Grbic et~al\mbox{.}(2021)]%
        {grbic_2021_evocraft}
\bibfield{author}{\bibinfo{person}{Djordje Grbic}, \bibinfo{person}{Rasmus~Berg
  Palm}, \bibinfo{person}{Elias Najarro}, \bibinfo{person}{Claire Glanois},
  {and} \bibinfo{person}{Sebastian Risi}.} \bibinfo{year}{2021}\natexlab{}.
\newblock \showarticletitle{{EvoCraft}: A new challenge for open-endedness}. In
  \bibinfo{booktitle}{\emph{Proceedings of the European Conference on
  Applications of Evolutionary Computation}}. \bibinfo{publisher}{Springer}.
\newblock


\bibitem[Guo et~al\mbox{.}(2025)]%
        {guo2025deepseek}
\bibfield{author}{\bibinfo{person}{Daya Guo}, \bibinfo{person}{Dejian Yang},
  \bibinfo{person}{Haowei Zhang}, \bibinfo{person}{Junxiao Song},
  \bibinfo{person}{Ruoyu Zhang}, \bibinfo{person}{Runxin Xu},
  \bibinfo{person}{Qihao Zhu}, \bibinfo{person}{Shirong Ma},
  \bibinfo{person}{Peiyi Wang}, \bibinfo{person}{Xiao Bi}, {et~al\mbox{.}}}
  \bibinfo{year}{2025}\natexlab{}.
\newblock \showarticletitle{Deepseek-r1: Incentivizing reasoning capability in
  llms via reinforcement learning}.
\newblock \bibinfo{journal}{\emph{arXiv preprint arXiv:2501.12948}}
  (\bibinfo{year}{2025}).
\newblock


\bibitem[Guzdial et~al\mbox{.}(2018)]%
        {guzdial2019benchmark}
\bibfield{author}{\bibinfo{person}{Matthew Guzdial}, \bibinfo{person}{Nicholas
  Liao}, \bibinfo{person}{Vishwa Shah}, {and} \bibinfo{person}{Mark~O Riedl}.}
  \bibinfo{year}{2018}\natexlab{}.
\newblock \showarticletitle{Creative Invention Benchmark}. In
  \bibinfo{booktitle}{\emph{Proceedings of the International Conference on
  Computational Creativity}}. \bibinfo{publisher}{ICCC}.
\newblock


\bibitem[Guzdial and Riedl(2016)]%
        {guzdial_2016_toward}
\bibfield{author}{\bibinfo{person}{Matthew Guzdial} {and} \bibinfo{person}{Mark
  Riedl}.} \bibinfo{year}{2016}\natexlab{}.
\newblock \showarticletitle{Toward game level generation from gameplay videos}.
  In \bibinfo{booktitle}{\emph{Proceedings of the Foundations of Digital Games
  Conference}}. \bibinfo{publisher}{ACM}.
\newblock


\bibitem[Hastings et~al\mbox{.}(2009)]%
        {hastings_2009_automatic}
\bibfield{author}{\bibinfo{person}{Erin~Jonathan Hastings},
  \bibinfo{person}{Ratan~K Guha}, {and} \bibinfo{person}{Kenneth~O Stanley}.}
  \bibinfo{year}{2009}\natexlab{}.
\newblock \showarticletitle{Automatic content generation in the {Galactic Arms
  Race} video game}.
\newblock \bibinfo{journal}{\emph{Transactions on Computational Intelligence
  and AI in Games}} \bibinfo{volume}{1}, \bibinfo{number}{4}
  (\bibinfo{year}{2009}).
\newblock


\bibitem[Horn et~al\mbox{.}(2014)]%
        {horn_2014_comparative}
\bibfield{author}{\bibinfo{person}{Britton Horn}, \bibinfo{person}{Steve
  Dahlskog}, \bibinfo{person}{Noor Shaker}, \bibinfo{person}{Gillian Smith},
  {and} \bibinfo{person}{Julian Togelius}.} \bibinfo{year}{2014}\natexlab{}.
\newblock \showarticletitle{A comparative evaluation of procedural level
  generators in the mario ai framework}. In
  \bibinfo{booktitle}{\emph{Proceedings of the Foundations of Digital Games
  Conference}}. \bibinfo{publisher}{ACM}.
\newblock


\bibitem[Jiang et~al\mbox{.}(2022)]%
        {jiang_2022_learning}
\bibfield{author}{\bibinfo{person}{Zehua Jiang}, \bibinfo{person}{Sam Earle},
  \bibinfo{person}{Michael Green}, {and} \bibinfo{person}{Julian Togelius}.}
  \bibinfo{year}{2022}\natexlab{}.
\newblock \showarticletitle{Learning controllable 3D level generators}. In
  \bibinfo{booktitle}{\emph{Proceedings of the Foundations of Digital Games
  Conference}}. \bibinfo{publisher}{ACM}.
\newblock


\bibitem[Karakovskiy and Togelius(2012)]%
        {karakovskiy_2012_mario}
\bibfield{author}{\bibinfo{person}{Sergey Karakovskiy} {and}
  \bibinfo{person}{Julian Togelius}.} \bibinfo{year}{2012}\natexlab{}.
\newblock \showarticletitle{The {Mario AI} benchmark and competitions}.
\newblock \bibinfo{journal}{\emph{Transactions on Computational Intelligence
  and AI in Games}} \bibinfo{volume}{4}, \bibinfo{number}{1}
  (\bibinfo{year}{2012}).
\newblock


\bibitem[Karth and Smith(2017)]%
        {karth2017wavefunctioncollapse}
\bibfield{author}{\bibinfo{person}{Isaac Karth} {and} \bibinfo{person}{Adam~M
  Smith}.} \bibinfo{year}{2017}\natexlab{}.
\newblock \showarticletitle{WaveFunctionCollapse is constraint solving in the
  wild}. In \bibinfo{booktitle}{\emph{Proceedings of the Foundations of Digital
  Games Conference}}. \bibinfo{publisher}{ACM}.
\newblock


\bibitem[Khalifa et~al\mbox{.}(2020)]%
        {khalifa_2020_pcgrl}
\bibfield{author}{\bibinfo{person}{Ahmed Khalifa}, \bibinfo{person}{Philip
  Bontrager}, \bibinfo{person}{Sam Earle}, {and} \bibinfo{person}{Julian
  Togelius}.} \bibinfo{year}{2020}\natexlab{}.
\newblock \showarticletitle{Pcgrl: Procedural content generation via
  reinforcement learning}. In \bibinfo{booktitle}{\emph{Proceedings of the
  Artificial Intelligence and Interactive Digital Entertainment Conference}}.
  \bibinfo{publisher}{AAAI}.
\newblock


\bibitem[Khalifa et~al\mbox{.}(2019)]%
        {khalifa_2019_elimination}
\bibfield{author}{\bibinfo{person}{Ahmed Khalifa}, \bibinfo{person}{Dan
  Gopstein}, {and} \bibinfo{person}{Julian Togelius}.}
  \bibinfo{year}{2019}\natexlab{}.
\newblock \showarticletitle{{ELIMINATION} from Design to Analysis}. In
  \bibinfo{booktitle}{\emph{Proceedings of the Conference on Games}}.
  \bibinfo{publisher}{IEEE}.
\newblock


\bibitem[Khalifa et~al\mbox{.}(2018a)]%
        {khalifa2018talakat}
\bibfield{author}{\bibinfo{person}{Ahmed Khalifa}, \bibinfo{person}{Scott Lee},
  \bibinfo{person}{Andy Nealen}, {and} \bibinfo{person}{Julian Togelius}.}
  \bibinfo{year}{2018}\natexlab{a}.
\newblock \showarticletitle{Talakat: Bullet hell generation through constrained
  {MAP-Elites}}. In \bibinfo{booktitle}{\emph{Proceedings of the Genetic and
  Evolutionary Computation Conference}}. \bibinfo{publisher}{ACM}.
\newblock


\bibitem[Khalifa et~al\mbox{.}(2018b)]%
        {khalifa_2018_talakat}
\bibfield{author}{\bibinfo{person}{Ahmed Khalifa}, \bibinfo{person}{Scott Lee},
  \bibinfo{person}{Andy Nealen}, {and} \bibinfo{person}{Julian Togelius}.}
  \bibinfo{year}{2018}\natexlab{b}.
\newblock \showarticletitle{Talakat: Bullet hell generation through constrained
  {MAP-Elites}}. In \bibinfo{booktitle}{\emph{Proceedings of the Genetic and
  Evolutionary Computation Conference}}. \bibinfo{publisher}{ACM}.
\newblock


\bibitem[Khalifa et~al\mbox{.}(2016)]%
        {khalifa2016general}
\bibfield{author}{\bibinfo{person}{Ahmed Khalifa}, \bibinfo{person}{Diego
  Perez-Liebana}, \bibinfo{person}{Simon~M Lucas}, {and}
  \bibinfo{person}{Julian Togelius}.} \bibinfo{year}{2016}\natexlab{}.
\newblock \showarticletitle{General video game level generation}. In
  \bibinfo{booktitle}{\emph{Proceedings of the Genetic and Evolutionary
  Computation Conference}}. \bibinfo{publisher}{ACM}.
\newblock


\bibitem[Kim et~al\mbox{.}(2019)]%
        {kim_2019_automatic}
\bibfield{author}{\bibinfo{person}{Hwanhee Kim}, \bibinfo{person}{Seongtaek
  Lee}, \bibinfo{person}{Hyundong Lee}, \bibinfo{person}{Teasung Hahn}, {and}
  \bibinfo{person}{Shinjin Kang}.} \bibinfo{year}{2019}\natexlab{}.
\newblock \showarticletitle{Automatic generation of game content using a
  graph-based wave function collapse algorithm}. In
  \bibinfo{booktitle}{\emph{Proceedings of the Conference on Games}}.
  \bibinfo{publisher}{IEEE}.
\newblock


\bibitem[Koster(2013)]%
        {koster_2013_theory}
\bibfield{author}{\bibinfo{person}{Raph Koster}.}
  \bibinfo{year}{2013}\natexlab{}.
\newblock \bibinfo{booktitle}{\emph{Theory of fun for game design}}.
\newblock \bibinfo{publisher}{O'Reilly Media, Inc}.
\newblock


\bibitem[Liapis(2016)]%
        {liapis2016arcade}
\bibfield{author}{\bibinfo{person}{Antonios Liapis}.}
  \bibinfo{year}{2016}\natexlab{}.
\newblock \showarticletitle{Exploring the Visual Styles of Arcade Game Assets}.
  In \bibinfo{booktitle}{\emph{Proceedings of Evolutionary and Biologically
  Inspired Music, Sound, Art and Design (EvoMusArt)}}.
  \bibinfo{publisher}{Springer}.
\newblock


\bibitem[Liapis(2020)]%
        {liapis2020tenyears}
\bibfield{author}{\bibinfo{person}{Antonios Liapis}.}
  \bibinfo{year}{2020}\natexlab{}.
\newblock \showarticletitle{10 Years of the {PCG} workshop: Past and Future
  Trends}. In \bibinfo{booktitle}{\emph{Proceedings of the FDG Workshop on
  Procedural Content Generation}}. \bibinfo{publisher}{ACM}.
\newblock


\bibitem[Liapis et~al\mbox{.}(2015)]%
        {liapis_2015_procedural}
\bibfield{author}{\bibinfo{person}{Antonios Liapis},
  \bibinfo{person}{Christoffer Holmg{\aa}rd}, \bibinfo{person}{Georgios~N
  Yannakakis}, {and} \bibinfo{person}{Julian Togelius}.}
  \bibinfo{year}{2015}\natexlab{}.
\newblock \showarticletitle{Procedural personas as critics for dungeon
  generation}. In \bibinfo{booktitle}{\emph{Proceedings of the European
  Conference on Applications of Evolutionary Computation}}.
  \bibinfo{publisher}{Springer}.
\newblock


\bibitem[Liapis et~al\mbox{.}(2013a)]%
        {liapis_2013_transforming}
\bibfield{author}{\bibinfo{person}{Antonios Liapis},
  \bibinfo{person}{H{\'e}ctor~P Mart{\'\i}nez}, \bibinfo{person}{Julian
  Togelius}, {and} \bibinfo{person}{Georgios~N Yannakakis}.}
  \bibinfo{year}{2013}\natexlab{a}.
\newblock \showarticletitle{Transforming exploratory creativity with DeLeNoX}.
  In \bibinfo{booktitle}{\emph{International Conference on Computational
  Creativity}}. \bibinfo{publisher}{ICCC}.
\newblock


\bibitem[Liapis et~al\mbox{.}(2013b)]%
        {liapis2013sentient}
\bibfield{author}{\bibinfo{person}{Antonios Liapis},
  \bibinfo{person}{Georgios~N Yannakakis}, {and} \bibinfo{person}{Julian
  Togelius}.} \bibinfo{year}{2013}\natexlab{b}.
\newblock \showarticletitle{Sentient sketchbook: computer-assisted game level
  authoring}. In \bibinfo{booktitle}{\emph{Proceedings of the Foundations of
  Digital Games Conference}}. \bibinfo{publisher}{ACM}.
\newblock


\bibitem[Lopes et~al\mbox{.}(2015)]%
        {lopes_2015_targeting}
\bibfield{author}{\bibinfo{person}{Phil Lopes}, \bibinfo{person}{Antonios
  Liapis}, {and} \bibinfo{person}{Georgios Yannakakis}.}
  \bibinfo{year}{2015}\natexlab{}.
\newblock \showarticletitle{Targeting horror via level and soundscape
  generation}. In \bibinfo{booktitle}{\emph{Proceedings of the Artificial
  Intelligence and Interactive Digital Entertainment Conference}}.
  \bibinfo{publisher}{AAAI}.
\newblock


\bibitem[Mnih et~al\mbox{.}(2013)]%
        {mnih_2013_playing}
\bibfield{author}{\bibinfo{person}{Volodymyr Mnih}, \bibinfo{person}{Koray
  Kavukcuoglu}, \bibinfo{person}{David Silver}, \bibinfo{person}{Alex Graves},
  \bibinfo{person}{Ioannis Antonoglou}, \bibinfo{person}{Daan Wierstra}, {and}
  \bibinfo{person}{Martin Riedmiller}.} \bibinfo{year}{2013}\natexlab{}.
\newblock \showarticletitle{Playing Atari with Deep Reinforcement Learning}.
\newblock \bibinfo{journal}{\emph{arXiv preprint arXiv:1312.5602}}
  \bibinfo{volume}{abs/1312.5602} (\bibinfo{year}{2013}).
\newblock


\bibitem[Mouret and Clune(2015)]%
        {mouret2015illuminating}
\bibfield{author}{\bibinfo{person}{Jean-Baptiste Mouret} {and}
  \bibinfo{person}{Jeff Clune}.} \bibinfo{year}{2015}\natexlab{}.
\newblock \showarticletitle{Illuminating search spaces by mapping elites}.
\newblock \bibinfo{journal}{\emph{arXiv preprint arXiv:1504.04909}}
  \bibinfo{volume}{abs/1504.04909} (\bibinfo{year}{2015}).
\newblock


\bibitem[OpenAI(2024)]%
        {openai2024reasoning}
\bibfield{author}{\bibinfo{person}{OpenAI}.} \bibinfo{year}{2024}\natexlab{}.
\newblock \bibinfo{title}{Learning to reason with {LLM}s}.
\newblock
  \bibinfo{howpublished}{\url{https://openai.com/index/learning-to-reason-with-llms/}}.
\newblock
\newblock
\shownote{Accessed On: 24-03-2025}.


\bibitem[Perez-Liebana et~al\mbox{.}(2019)]%
        {perez_2019_general}
\bibfield{author}{\bibinfo{person}{Diego Perez-Liebana},
  \bibinfo{person}{Jialin Liu}, \bibinfo{person}{Ahmed Khalifa},
  \bibinfo{person}{Raluca~D Gaina}, \bibinfo{person}{Julian Togelius}, {and}
  \bibinfo{person}{Simon~M Lucas}.} \bibinfo{year}{2019}\natexlab{}.
\newblock \showarticletitle{General video game ai: A multitrack framework for
  evaluating agents, games, and content generation algorithms}.
\newblock \bibinfo{journal}{\emph{Transactions on Games}} \bibinfo{volume}{11},
  \bibinfo{number}{3} (\bibinfo{year}{2019}).
\newblock


\bibitem[Pugh et~al\mbox{.}(2016)]%
        {pugh2016qd}
\bibfield{author}{\bibinfo{person}{Justin~K. Pugh}, \bibinfo{person}{Lisa~B.
  Soros}, {and} \bibinfo{person}{Kenneth~O. Stanley}.}
  \bibinfo{year}{2016}\natexlab{}.
\newblock \showarticletitle{Quality Diversity: A New Frontier for Evolutionary
  Computation}.
\newblock \bibinfo{journal}{\emph{Frontiers in Robotics and AI}}
  \bibinfo{volume}{3} (\bibinfo{year}{2016}).
\newblock


\bibitem[Ritchie(2007)]%
        {ritchie2007empirical}
\bibfield{author}{\bibinfo{person}{Graeme Ritchie}.}
  \bibinfo{year}{2007}\natexlab{}.
\newblock \showarticletitle{Some Empirical Criteria for Attributing Creativity
  to a Computer Program}.
\newblock \bibinfo{journal}{\emph{Minds and Machines}} \bibinfo{volume}{17},
  \bibinfo{number}{1} (\bibinfo{year}{2007}).
\newblock


\bibitem[Salge et~al\mbox{.}(2018)]%
        {salge_2018_generative}
\bibfield{author}{\bibinfo{person}{Christoph Salge},
  \bibinfo{person}{Michael~Cerny Green}, \bibinfo{person}{Rodgrigo Canaan},
  {and} \bibinfo{person}{Julian Togelius}.} \bibinfo{year}{2018}\natexlab{}.
\newblock \showarticletitle{Generative design in minecraft (gdmc) settlement
  generation competition}. In \bibinfo{booktitle}{\emph{Proceedings of the
  Foundations of Digital Games Conference}}. \bibinfo{publisher}{ACM}.
\newblock


\bibitem[Shaker et~al\mbox{.}(2016)]%
        {shaker2016constructive}
\bibfield{author}{\bibinfo{person}{Noor Shaker}, \bibinfo{person}{Antonios
  Liapis}, \bibinfo{person}{Julian Togelius}, \bibinfo{person}{Ricardo Lopes},
  {and} \bibinfo{person}{Rafael Bidarra}.} \bibinfo{year}{2016}\natexlab{}.
\newblock \showarticletitle{Constructive Generation Methods for Dungeons and
  Levels}.
\newblock In \bibinfo{booktitle}{\emph{Procedural Content Generation in Games:
  A Textbook and an Overview of Current Research}},
  \bibfield{editor}{\bibinfo{person}{Noor Shaker}, \bibinfo{person}{Julian
  Togelius}, {and} \bibinfo{person}{Mark~J. Nelson}} (Eds.).
  \bibinfo{publisher}{Springer}, \bibinfo{pages}{31--55}.
\newblock


\bibitem[Shaker et~al\mbox{.}(2012)]%
        {shaker2012evolving}
\bibfield{author}{\bibinfo{person}{Noor Shaker}, \bibinfo{person}{Miguel
  Nicolau}, \bibinfo{person}{Georgios~N Yannakakis}, \bibinfo{person}{Julian
  Togelius}, {and} \bibinfo{person}{Michael O'neill}.}
  \bibinfo{year}{2012}\natexlab{}.
\newblock \showarticletitle{Evolving levels for super mario bros using
  grammatical evolution}. In \bibinfo{booktitle}{\emph{Proceedings of the
  Conference on Computational Intelligence and Games}}.
  \bibinfo{publisher}{IEEE}, \bibinfo{pages}{304--311}.
\newblock


\bibitem[Silver et~al\mbox{.}(2016)]%
        {silver_2016_mastering}
\bibfield{author}{\bibinfo{person}{David Silver}, \bibinfo{person}{Aja Huang},
  \bibinfo{person}{Chris~J Maddison}, \bibinfo{person}{Arthur Guez},
  \bibinfo{person}{Laurent Sifre}, \bibinfo{person}{George Van Den~Driessche},
  \bibinfo{person}{Julian Schrittwieser}, \bibinfo{person}{Ioannis Antonoglou},
  \bibinfo{person}{Veda Panneershelvam}, \bibinfo{person}{Marc Lanctot},
  {et~al\mbox{.}}} \bibinfo{year}{2016}\natexlab{}.
\newblock \showarticletitle{Mastering the game of Go with deep neural networks
  and tree search}.
\newblock \bibinfo{journal}{\emph{Nature}}  \bibinfo{volume}{529}
  (\bibinfo{year}{2016}).
\newblock


\bibitem[Siper et~al\mbox{.}(2022)]%
        {siper_2022_path}
\bibfield{author}{\bibinfo{person}{Matthew Siper}, \bibinfo{person}{Ahmed
  Khalifa}, {and} \bibinfo{person}{Julian Togelius}.}
  \bibinfo{year}{2022}\natexlab{}.
\newblock \showarticletitle{Path of destruction: Learning an iterative level
  generator using a small dataset}. In \bibinfo{booktitle}{\emph{Proceedings of
  the Symposium Series on Computational Intelligence}}.
  \bibinfo{publisher}{IEEE}.
\newblock


\bibitem[Summerville et~al\mbox{.}(2018)]%
        {summerville_2018_procedural}
\bibfield{author}{\bibinfo{person}{Adam Summerville}, \bibinfo{person}{Sam
  Snodgrass}, \bibinfo{person}{Matthew Guzdial}, \bibinfo{person}{Christoffer
  Holmg{\aa}rd}, \bibinfo{person}{Amy~K Hoover}, \bibinfo{person}{Aaron
  Isaksen}, \bibinfo{person}{Andy Nealen}, {and} \bibinfo{person}{Julian
  Togelius}.} \bibinfo{year}{2018}\natexlab{}.
\newblock \showarticletitle{Procedural content generation via machine learning
  {(PCGML)}}.
\newblock \bibinfo{journal}{\emph{Transactions on Games}} \bibinfo{volume}{10},
  \bibinfo{number}{3} (\bibinfo{year}{2018}).
\newblock


\bibitem[Togelius and Schmidhuber(2008)]%
        {togelius_2008_experiment}
\bibfield{author}{\bibinfo{person}{Julian Togelius} {and}
  \bibinfo{person}{Jurgen Schmidhuber}.} \bibinfo{year}{2008}\natexlab{}.
\newblock \showarticletitle{An experiment in automatic game design}. In
  \bibinfo{booktitle}{\emph{Proceedings of the Symposium On Computational
  Intelligence and Games}}. \bibinfo{publisher}{IEEE}.
\newblock


\bibitem[Togelius et~al\mbox{.}(2011)]%
        {togelius_2011_search}
\bibfield{author}{\bibinfo{person}{Julian Togelius},
  \bibinfo{person}{Georgios~N Yannakakis}, \bibinfo{person}{Kenneth~O Stanley},
  {and} \bibinfo{person}{Cameron Browne}.} \bibinfo{year}{2011}\natexlab{}.
\newblock \showarticletitle{Search-based procedural content generation: A
  taxonomy and survey}.
\newblock \bibinfo{journal}{\emph{Transactions on Computational Intelligence
  and AI in Games}} \bibinfo{volume}{3}, \bibinfo{number}{3}
  (\bibinfo{year}{2011}).
\newblock


\bibitem[Torrado et~al\mbox{.}(2020)]%
        {torrado_2020_bootstrapping}
\bibfield{author}{\bibinfo{person}{Ruben~Rodriguez Torrado},
  \bibinfo{person}{Ahmed Khalifa}, \bibinfo{person}{Michael~Cerny Green},
  \bibinfo{person}{Niels Justesen}, \bibinfo{person}{Sebastian Risi}, {and}
  \bibinfo{person}{Julian Togelius}.} \bibinfo{year}{2020}\natexlab{}.
\newblock \showarticletitle{Bootstrapping conditional {GANs} for video game
  level generation}. In \bibinfo{booktitle}{\emph{Proceedings of the Conference
  on Games}}. \bibinfo{publisher}{IEEE}.
\newblock


\bibitem[Vinyals et~al\mbox{.}(2019)]%
        {vinyals_2019_grandmaster}
\bibfield{author}{\bibinfo{person}{Oriol Vinyals}, \bibinfo{person}{Igor
  Babuschkin}, \bibinfo{person}{Wojciech~M Czarnecki},
  \bibinfo{person}{Micha{\"e}l Mathieu}, \bibinfo{person}{Andrew Dudzik},
  \bibinfo{person}{Junyoung Chung}, \bibinfo{person}{David~H Choi},
  \bibinfo{person}{Richard Powell}, \bibinfo{person}{Timo Ewalds},
  \bibinfo{person}{Petko Georgiev}, {et~al\mbox{.}}}
  \bibinfo{year}{2019}\natexlab{}.
\newblock \showarticletitle{Grandmaster level in StarCraft II using multi-agent
  reinforcement learning}.
\newblock \bibinfo{journal}{\emph{Nature}}  \bibinfo{volume}{575}
  (\bibinfo{year}{2019}).
\newblock


\bibitem[Volz et~al\mbox{.}(2018)]%
        {volz_2018_evolving}
\bibfield{author}{\bibinfo{person}{Vanessa Volz}, \bibinfo{person}{Jacob
  Schrum}, \bibinfo{person}{Jialin Liu}, \bibinfo{person}{Simon~M Lucas},
  \bibinfo{person}{Adam Smith}, {and} \bibinfo{person}{Sebastian Risi}.}
  \bibinfo{year}{2018}\natexlab{}.
\newblock \showarticletitle{Evolving {Mario} levels in the latent space of a
  deep convolutional generative adversarial network}. In
  \bibinfo{booktitle}{\emph{Proceedings of the Genetic and evolutionary
  computation conference}}. \bibinfo{publisher}{ACM}.
\newblock


\bibitem[Wang et~al\mbox{.}(2022)]%
        {wang_2022_fun}
\bibfield{author}{\bibinfo{person}{Ziqi Wang}, \bibinfo{person}{Jialin Liu},
  {and} \bibinfo{person}{Georgios~N Yannakakis}.}
  \bibinfo{year}{2022}\natexlab{}.
\newblock \showarticletitle{The fun facets of {Mario}: Multifaceted
  experience-driven PCG via reinforcement learning}. In
  \bibinfo{booktitle}{\emph{Proceedings of the Foundations of Digital Games
  Conference}}. \bibinfo{publisher}{ACM}.
\newblock


\bibitem[Yannakakis and Togelius(2011)]%
        {yannakakis_2011_experience}
\bibfield{author}{\bibinfo{person}{Georgios~N Yannakakis} {and}
  \bibinfo{person}{Julian Togelius}.} \bibinfo{year}{2011}\natexlab{}.
\newblock \showarticletitle{Experience-driven procedural content generation}.
\newblock \bibinfo{journal}{\emph{Transactions on Affective Computing}}
  \bibinfo{volume}{2}, \bibinfo{number}{3} (\bibinfo{year}{2011}).
\newblock


\bibitem[Yannakakis and Togelius(2014)]%
        {yannakakis_2014_panorama}
\bibfield{author}{\bibinfo{person}{Georgios~N Yannakakis} {and}
  \bibinfo{person}{Julian Togelius}.} \bibinfo{year}{2014}\natexlab{}.
\newblock \showarticletitle{A panorama of artificial and computational
  intelligence in games}.
\newblock \bibinfo{journal}{\emph{Transactions on Computational Intelligence
  and AI in Games}} \bibinfo{volume}{7}, \bibinfo{number}{4}
  (\bibinfo{year}{2014}).
\newblock


\bibitem[Yao et~al\mbox{.}(2024)]%
        {yao2024survey}
\bibfield{author}{\bibinfo{person}{Yifan Yao}, \bibinfo{person}{Jinhao Duan},
  \bibinfo{person}{Kaidi Xu}, \bibinfo{person}{Yuanfang Cai},
  \bibinfo{person}{Zhibo Sun}, {and} \bibinfo{person}{Yue Zhang}.}
  \bibinfo{year}{2024}\natexlab{}.
\newblock \showarticletitle{A survey on large language model ({LLM}) security
  and privacy: The good, the bad, and the ugly}.
\newblock \bibinfo{journal}{\emph{High-Confidence Computing}}
  (\bibinfo{year}{2024}), \bibinfo{pages}{100211}.
\newblock


\end{thebibliography}

\begin{appendices}

\section{Experiments with LLM Generators}

With the surge of LLMs~\cite{yao2024survey}, we wanted to showcase that the PCG Benchmark can be used to compare LLM generators~\cite{gallotta2024large} to more classic approaches. We compared 100 separate runs of the provided algorithms (Random, ES, and GA) with a constructive generator~\cite{shaker2016constructive} and two few-shot LLM generators based on \textit{Llama 3.2} and \textit{DeepSeek-r1}. We tested them on only 3 simple problems of the PCG Benchmark: Binary, Sokoban, and Zelda. The constructive generator starts by building a 2D maze using Prim's algorithm~\cite{buck2015mazes}. This generated maze is used as-is for Binary. For Sokoban and Zelda, the script erases more than 50\% of the walls to allow for open areas, then adds the missing objects in the level at random locations (for Sokoban, these locations are restricted such that crates have no more than one side blocked). LLM generators use a simple prompt that explains the goal of the game and how to play it, followed by the goal of the generator and five example levels.

\begin{figure}[t]
    \centering
    \includegraphics[width=\linewidth]{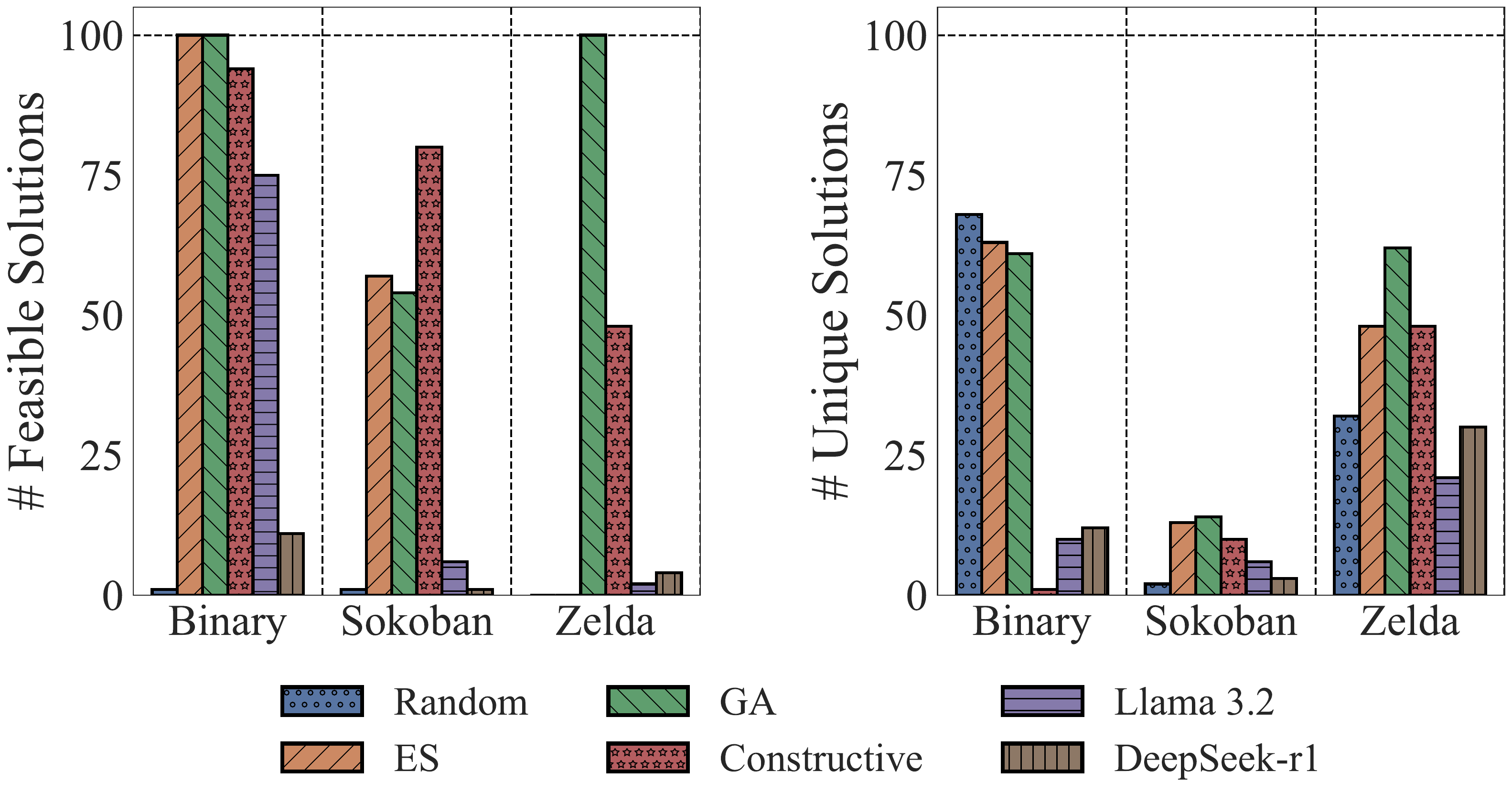}
    \caption{The number of feasible and unique solutions over 100 separate runs on Binary, Sokoban, and Zelda using six different methods (three search-based generators, one constructive generator, and two few-shot LLM generators).}
    \label{fig:appendix_results}
\end{figure}

Figure~\ref{fig:appendix_results} shows the comparison between these algorithms from the perspective of quality (number of feasible solutions) and diversity (number of unique solutions) over 100 runs, extending the findings from Fig.~\ref{fig:success_graphs}. GA has more feasible solutions overall, although the constructive algorithm surpasses it in Sokoban. In Sokoban, the script is fairly thorough (e.g. constraining where crates can be placed) and thus it is not surprising that it can generate many feasible solutions. It is worth noting, however, that most of these solutions are not unique: GA and ES find more unique feasible results even if the number of feasible results is fewer than for the constructive method. From LLM methods, \textit{Llama 3.2} outperforms \textit{DeepSeek-r1} in terms of quality for Binary and Sokoban, but both LLMs find very few feasible solutions in Zelda. This disparity was surprising because reasoning models such as \textit{DeepSeek-r1} usually perform better on language tasks than traditional models such as \textit{Llama 3.2}~\cite{openai2024reasoning}. Looking at the generated examples, we noticed that \textit{Llama 3.2} has a higher chance of copying some of the examples used in the prompt. It seems that the added reasoning tokens~\cite{guo2025deepseek} in \textit{DeepSeek-r1} might have pushed it towards understanding the examples in the prompt and trying to create new ones; however, its results were often infeasible. This first, preliminary experiment showcases how the PCG Benchmark can be used for current and emerging technologies such as LLM-based generation~\cite{gallotta2024large}.

\end{appendices}

\end{document}